\documentclass{article}

\PassOptionsToPackage{table}{xcolor}
\usepackage{report}

\usepackage[utf8]{inputenc} 
\usepackage[T1]{fontenc}    
\usepackage{hyperref}       
\usepackage{url}            
\usepackage{booktabs}       
\usepackage{amsfonts}       
\usepackage{nicefrac}       
\usepackage{microtype}      
\usepackage{xcolor}         

\usepackage{amsthm}

\usepackage{amsmath,amssymb,amsfonts}
\usepackage{booktabs}
\usepackage{tikz}

\usepackage{wrapfig} 

\usepackage{algorithm}
\usepackage{graphicx}
\usepackage{float}
\usepackage{tikz}
\usetikzlibrary{arrows.meta,positioning,calc}

\usepackage{algpseudocode}

\usepackage{bbm}
\usepackage{booktabs}
\usepackage{multirow}
\usepackage{makecell}
\usepackage{xcolor}
\usepackage{graphicx}
\usepackage[table]{xcolor}

\definecolor{bestred}{RGB}{190,30,45}
\definecolor{secondblue}{RGB}{35,95,180}
\definecolor{thirdgreen}{RGB}{35,140,95}
\newcommand{\best}[1]{\textcolor{bestred}{\textbf{#1}}}
\newcommand{\second}[1]{\textcolor{secondblue}{\textbf{#1}}}
\newcommand{\third}[1]{\textcolor{thirdgreen}{\textbf{#1}}}

\definecolor{rankBest}{RGB}{190, 38, 45}
\definecolor{rankSecond}{RGB}{35, 92, 175}
\definecolor{rankThird}{RGB}{30, 125, 75}

\usepackage{amsmath,amssymb,amsfonts,bm}
\usepackage{graphicx}
\usepackage{tikz}
\usetikzlibrary{arrows.meta,positioning,calc,fit,backgrounds}

\newcommand{\NA}{\textemdash}

\definecolor{groupgray}{RGB}{238,238,238}

\definecolor{inactivegray}{RGB}{120,120,120}

\definecolor{grouprow}{RGB}{238,238,238}

\usepackage{tabularx}
\usepackage{array}
\newcolumntype{Y}{>{\raggedright\arraybackslash}X}

\newcommand{\abldelta}[1]{\textcolor{secondblue}{\scriptsize\,#1}}

\usepackage{tabularx}
\usepackage{array}
\usepackage[table]{xcolor}

\definecolor{appBlue}{RGB}{34, 88, 153}
\definecolor{appDark}{RGB}{45, 50, 62}
\definecolor{appGray}{RGB}{92, 99, 112}
\definecolor{appLight}{RGB}{246, 249, 254}
\definecolor{appLine}{RGB}{216, 224, 236}

\definecolor{ablationblue}{RGB}{35,95,180}

\newcolumntype{A}{>{\raggedright\arraybackslash}p{0.30\linewidth}}
\newcolumntype{B}{>{\raggedright\arraybackslash}X}

\usepackage{wrapfig}

\newcommand{\appSection}[2]{%
\hyperref[#1]{%
\textbf{\textcolor{appBlue}{Appendix~\ref*{#1}}}}%
\par
\vspace{1pt}
\hyperref[#1]{%
\textbf{\textcolor{appDark}{#2}}}%
}

\newcommand{\appSub}[2]{%
\hyperref[#1]{%
\textcolor{appDark}{\ref*{#1}\quad #2}}%
\par
\vspace{2pt}
}

\usepackage[most]{tcolorbox}
\tcbuselibrary{breakable,skins}

\definecolor{njuPurple}{RGB}{220,205,230}
\definecolor{njuPurpleLight}{RGB}{250,245,252}

\newtcolorbox{abstractbox}{
    colback=njuPurpleLight,
    colframe=njuPurple,
    boxrule=1pt,
    arc=4mm,
    left=8pt,
    right=8pt,
    top=8pt,
    bottom=8pt,
    opacityback=0.95,
    breakable
}

\title{Hidden Forgetting in Continual Multimodal Learning: When Accuracy Survives but Grounding Fails}

\author{%
Qianyu Chen$^{1}$ \quad
Canran Xiao$^{2}$ \quad
Runxuan Tang$^{1}$\\[4pt]
$^{1}$Nanyang Technological University\\
$^{2}$Shenzhen Campus of Sun Yat-sen University
}

\begin{document}

\maketitle

\begin{abstract}
\begin{abstractbox}
Multimodal large language models must continually adapt to evolving tasks and domains, yet standard continual learning metrics mainly measure whether old answers remain correct, leaving the stability of multimodal grounding largely unexamined.
We study this overlooked failure mode and ask whether a continually adapted MLLM can preserve not only what it answers, but also how it uses visual, textual, OCR, chart, and document evidence.
We identify \emph{hidden evidence-use forgetting}, where answer accuracy is retained while the model silently shifts toward different or less grounded evidence channels, and propose \textsc{RCL}, a replay-free reliance-constrained continual learning framework.
\textsc{RCL} freezes the previous checkpoint as a behavioral reference, estimates teacher and student evidence-reliance profiles through counterfactual channel interventions, and jointly optimizes task learning, prediction preservation, and reliance preservation without adding inference-time cost.
Across CoIN, COAST, MCITlib, and an evidence-sensitive multimodal stream, \textsc{RCL} consistently improves final performance and reduces forgetting over replay-free, PEFT, routing, and memory-assisted baselines, while substantially lowering modality reliance drift, dominant evidence flips, and hidden forgetting rates.
These results suggest that robust continual multimodal learning requires preserving the evidence path behind correct answers, not merely the answers themselves.
\end{abstractbox}
\end{abstract}

\section{Introduction}
\label{sec:introduction}

Multimodal large language models (MLLMs) have become a central interface for visual question answering, document understanding, chart reasoning, embodied perception, and open-ended image--language interaction. In realistic deployments, however, these models rarely face a fixed task distribution: new domains, instructions, visual formats, and reasoning skills arrive sequentially, making continual adaptation essential. The key challenge is that improving an MLLM on new multimodal tasks can silently degrade previously acquired abilities, limiting its reliability in long-lived and safety-sensitive applications.

Recent studies have begun to formalize this problem as continual vision--language learning and continual multimodal instruction tuning. CLiMB highlights that vision--language systems suffer nontrivial forgetting across sequential multimodal tasks \citep{srinivasan2022climb}, while ZSCL and CTP show that continual adaptation can damage the zero-shot transfer and cross-modal topology of pretrained vision--language representations \citep{zheng2023zscl,zhu2023ctp}. More recent MLLM-oriented benchmarks, such as CoIN and Continual LLaVA, further demonstrate that instruction-tuned MLLMs are vulnerable to capability-level forgetting when exposed to evolving multimodal instruction streams \citep{chen2024coin,cao2024continual}. These works establish continual multimodal learning as an important problem and motivate a growing set of methods for mitigating forgetting.

Existing approaches have made substantial progress by preserving model parameters, output behavior, or task-specific adaptation modules. For example, Model Tailor mitigates forgetting by identifying sparse parameter modifications that preserve pretrained capabilities \citep{zhu2024model}, while HiDe-LLaVA, BranchLoRA, and ProgLoRA organize task-specific or progressive LoRA modules to reduce interference during continual instruction tuning \citep{guo2025hide_llava,zhang2025branchlora,yu2025proglora}. Other methods improve continual multimodal adaptation through prompt routing, expert allocation, or synthetic replay signals, as seen in ModalPrompt, CL-MoE, and GIFT \citep{zeng2025modalprompt,huai2025clmoe,wu2025gift}. SEFE further argues that forgetting in MLLMs should not be treated as a single phenomenon, distinguishing superficial response-format degradation from more essential knowledge loss \citep{chen2025sefe}. Despite these advances, most methods still evaluate and constrain forgetting mainly through final answers, output distributions, or parameter changes.

This answer-centric view leaves a critical gap. In multimodal reasoning, two checkpoints may produce the same answer while relying on different evidence sources, such as visual regions, OCR tokens, or language priors. Such hidden shifts are especially problematic because they can make a model appear stable under standard accuracy metrics while its grounding behavior becomes less faithful and more shortcut-driven. Prior work on language bias in VQA has shown that correct predictions may arise from dataset priors rather than genuine visual evidence \citep{agrawal2018dont}, and counterfactual analysis has been used to reveal causal effects of biased evidence in static VQA settings \citep{niu2021counterfactual}. However, this perspective has not been fully integrated into replay-free continual learning for MLLMs, where the central question is not only whether old answers are retained, but whether the model continues to use the appropriate multimodal evidence to obtain them.

In this paper, we ask: \emph{Can continual MLLM adaptation preserve not only what the model answers, but also how it uses multimodal evidence?} We address this question from the perspective of evidence-use stability. At a high level, our framework treats the previous model as a behavioral reference, estimates how both the previous and current models depend on available evidence channels, and constrains continual adaptation to preserve this reliance pattern while learning new tasks. This design complements existing output- or parameter-preservation strategies by targeting a hidden dimension of forgetting that standard answer-level evaluation can miss.

Our contributions are summarized as follows:
\begin{itemize}
    \item We identify and formalize \emph{hidden evidence-use forgetting}: an MLLM may retain the correct answer while drifting toward different or less grounded evidence sources during continual adaptation.
    \item We propose a replay-free reliance-preserving continual learning framework that constrains evidence-use drift without storing previous multimodal examples or increasing inference-time cost.
    \item Through continual multimodal instruction streams, we show that preserving reliance profiles improves both answer retention and evidence-grounded stability, providing a stronger notion of continual multimodal robustness than answer accuracy alone.
\end{itemize}

\section{Related Work}
\label{sec:related_work}

\paragraph{Continual vision--language representation learning.}
Early multimodal continual learning studies moved beyond unimodal classification by evaluating sequential vision--language tasks and cross-modal representation stability. CLiMB introduced a benchmark for continual vision-and-language tasks, while Mod-X, ZSCL, and CTP showed that sequential adaptation can distort intra-modal geometry, cross-modal alignment, and zero-shot transfer in CLIP-like models \citep{srinivasan2022climb,ni2023continual,zheng2023zscl,zhu2023ctp}. More recent work scales this direction to practical multimodal streams and stronger VLM backbones: FoMo-in-Flux studies continual multimodal pretraining under realistic compute and data-mixture constraints \citep{udandarao2024practitioner}; C-CLIP evaluates continual image--text retrieval and zero-shot retention \citep{liu2025c}; LADA introduces label-specific CLIP adapters \citep{luo2025lada}; and GIFT uses synthetic image--text data for replay-style preservation \citep{wu2025gift}. These methods substantially improve representation-level stability, but they mainly protect global embedding spaces, retrieval/classification performance, or pretrained zero-shot ability. In contrast, our work studies autoregressive MLLMs, where forgetting may occur not only in the final answer but also in the instance-specific allocation of reliance over visual regions, OCR tokens, chart elements, and language context.

\paragraph{Continual instruction tuning for MLLMs.}
Continual instruction tuning has recently become the dominant setting for adapting MLLMs to evolving tasks. CoIN formulates sequential instruction tuning over diverse task categories and shows that MLLMs suffer severe forgetting of instruction following \citep{chen2024coin}. Continual LLaVA further introduces COAST, covering domain-, capability-, and dataset-incremental LVLM adaptation \citep{cao2024continual}. Recent methods then focus on stronger parameter organization and routing: Model Tailor identifies sparse parameter patches to preserve pretrained capabilities \citep{zhu2024model}; HiDe-LLaVA uses task-specific expansion and task-general fusion based on layer-wise similarity \citep{guo2025hide_llava}; BranchLoRA and ProgLoRA improve LoRA allocation under MCIT \citep{zhang2025branchlora,yu2025proglora}; ModalPrompt uses dual-modality guided prompt fusion and selection \citep{zeng2025modalprompt}; CL-MoE introduces dual momentum experts for continual VQA \citep{huai2025clmoe}; and MLLM-CL proposes domain/ability continual settings with MLLM-based routing \citep{zhao2025mllm}. SEFE is particularly relevant because it separates superficial format forgetting from essential knowledge forgetting \citep{chen2025sefe}. However, these direct competitors primarily preserve answer accuracy, response format, parameter specialization, or output distributions. Our preliminary observation suggests that such preservation can still hide evidence-use drift. RCL therefore adds a complementary constraint: the previous checkpoint's counterfactual reliance profile serves as a teacher signal, preserving not only \emph{what} the model answers but also \emph{how} it uses multimodal evidence.

\paragraph{Grounding, shortcut bias, and counterfactual evidence use.}
A related line of VQA research shows that correct answers may arise from language priors rather than grounded visual evidence. VQA-CP exposes this issue by changing answer priors between training and testing \citep{agrawal2018dont}, and Counterfactual VQA uses causal counterfactual reasoning to estimate and reduce language-bias effects \citep{niu2021counterfactual}. These works motivate channel-level evidence diagnostics, but they are mainly designed for static debiasing or evaluation rather than continual adaptation. Our work brings this perspective into replay-free MLLM continual learning by turning counterfactual evidence suppression into a training-time reliance-preservation objective, while adding no inference-time overhead.

\section{Preliminary Study}
\label{sec:preliminary}

Before introducing our method, we ask whether continual multimodal adaptation forgets only final answers, or also the evidence path used to obtain them. 
This distinction matters because two checkpoints may produce the same answer while relying on different channels, such as visual regions, OCR tokens, chart elements, or language priors. 
Standard answer-level evaluation may therefore miss a form of \emph{hidden forgetting}: the answer is preserved, but the underlying multimodal evidence use has drifted.

\paragraph{Diagnostic setup.}
We conduct a replay-free pilot study on an eight-stage multimodal stream and evaluate two simple baselines: sequential LoRA tuning and output-level answer distillation. 
No component of our proposed method is used. 
After training, we perform post-hoc channel ablations to estimate a diagnostic evidence-use vector $\bm q(f;x,y)$ for each checkpoint $f$, where $q_m$ is proportional to the loss increase caused by suppressing evidence channel $m$. 
Full protocol details are provided in Appendix~\ref{app:prelim_details}.

Let $f_j$ be the checkpoint after learning stage $j$, and $f_k$ a later checkpoint after stage $k>j$. 
For an old evaluation set $\mathcal E_j$, we measure answer forgetting, evidence-sensitivity drift, and answer-preserved dominant-evidence flips as
\begin{equation}
\label{eq:prelim_metrics}
\begin{aligned}
A_{j\rightarrow k}
&=
{\rm Acc}(f_j;\mathcal E_j)
-
{\rm Acc}(f_k;\mathcal E_j),\\
G_{j\rightarrow k}
&=
\mathbb E_{(x,y)\sim\mathcal E_j}
\left[
\left\|
\bm q(f_j;x,y)-\bm q(f_k;x,y)
\right\|_1
\right],\\
H_{j\rightarrow k}
&=
\mathbb E
\left[
\mathbf{1}\!\left[
\arg\max_m q_m(f_j;x,y)
\neq
\arg\max_m q_m(f_k;x,y)
\right]
\,\middle|\,
f_j(x)=f_k(x)=y
\right].
\end{aligned}
\end{equation}
Here $A_{j\rightarrow k}$ captures visible answer degradation, whereas $G_{j\rightarrow k}$ and $H_{j\rightarrow k}$ diagnose whether the model still uses the same evidence when the answer is retained.

\begin{figure*}[t]
    \centering
    \includegraphics[width=\linewidth]{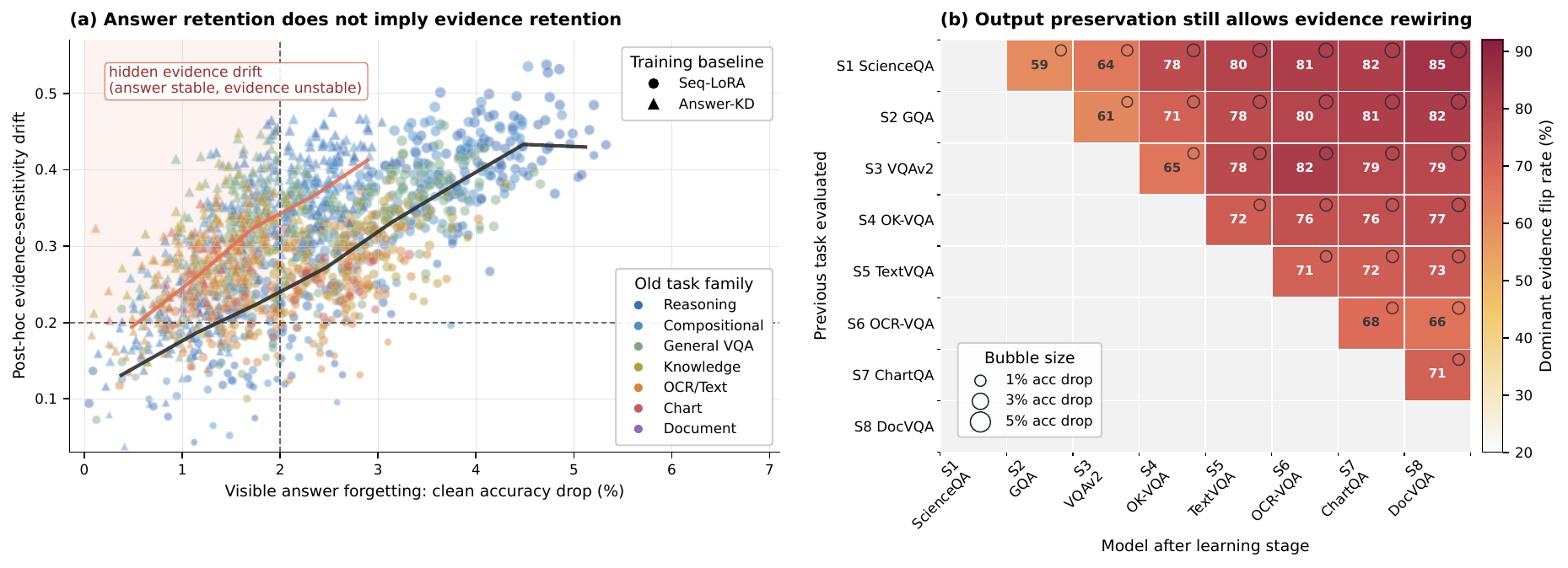}
    \caption{
    \textbf{Hidden evidence-use drift in replay-free multimodal continual adaptation.}
    \textbf{(a)} Answer forgetting is weakly aligned with post-hoc evidence drift; the shaded region marks small answer drop but large evidence drift.
    \textbf{(b)} For Answer-KD, warm cells with small bubbles show that output preservation can hide substantial dominant-evidence flips.
    Details are in Appendix~\ref{app:prelim_details}.
    }
    \label{fig:preliminary_hidden_forgetting}
\end{figure*}

\paragraph{Findings.}
Figure~\ref{fig:preliminary_hidden_forgetting} shows that answer preservation does not imply evidence preservation.
Many transitions incur less than $2\%$ clean accuracy drop while exhibiting large evidence-sensitivity drift.
Moreover, output-level distillation reduces the average clean accuracy drop from $2.72\%$ to $1.57\%$, but still leaves $66.3\%$ of diagnostic slices in the hidden-drift region, compared with $14.4\%$ for sequential LoRA.
The effect is most pronounced on OCR-, chart-, and document-centric stages, where the model often keeps the correct answer but shifts away from grounded visual/textual evidence toward easier priors.

These observations motivate the central principle of our method: continual multimodal learning should preserve not only \emph{what} the previous model answers, but also \emph{how} it uses evidence.
We next formalize this principle as reliance-constrained continual learning.

\section{Method}
\label{sec:method}

We propose \textbf{Reliance-Constrained Continual Learning} (RCL), a replay-free framework for mitigating hidden forgetting in continual multimodal learning. RCL is built on a simple principle: a continual MLLM should preserve not only its answers, but also the evidence channels that support those answers. At each learning stage, RCL freezes the previous model as a behavioral reference, probes both the reference and current models through counterfactual channel interventions, estimates their modality-reliance profiles, and updates the current model by jointly optimizing task learning, prediction preservation, and reliance preservation. The framework updates only parameter-efficient adaptation parameters and introduces no additional computation at inference time.

\begin{figure}[htb]
	\centering
	\includegraphics[width=\linewidth]{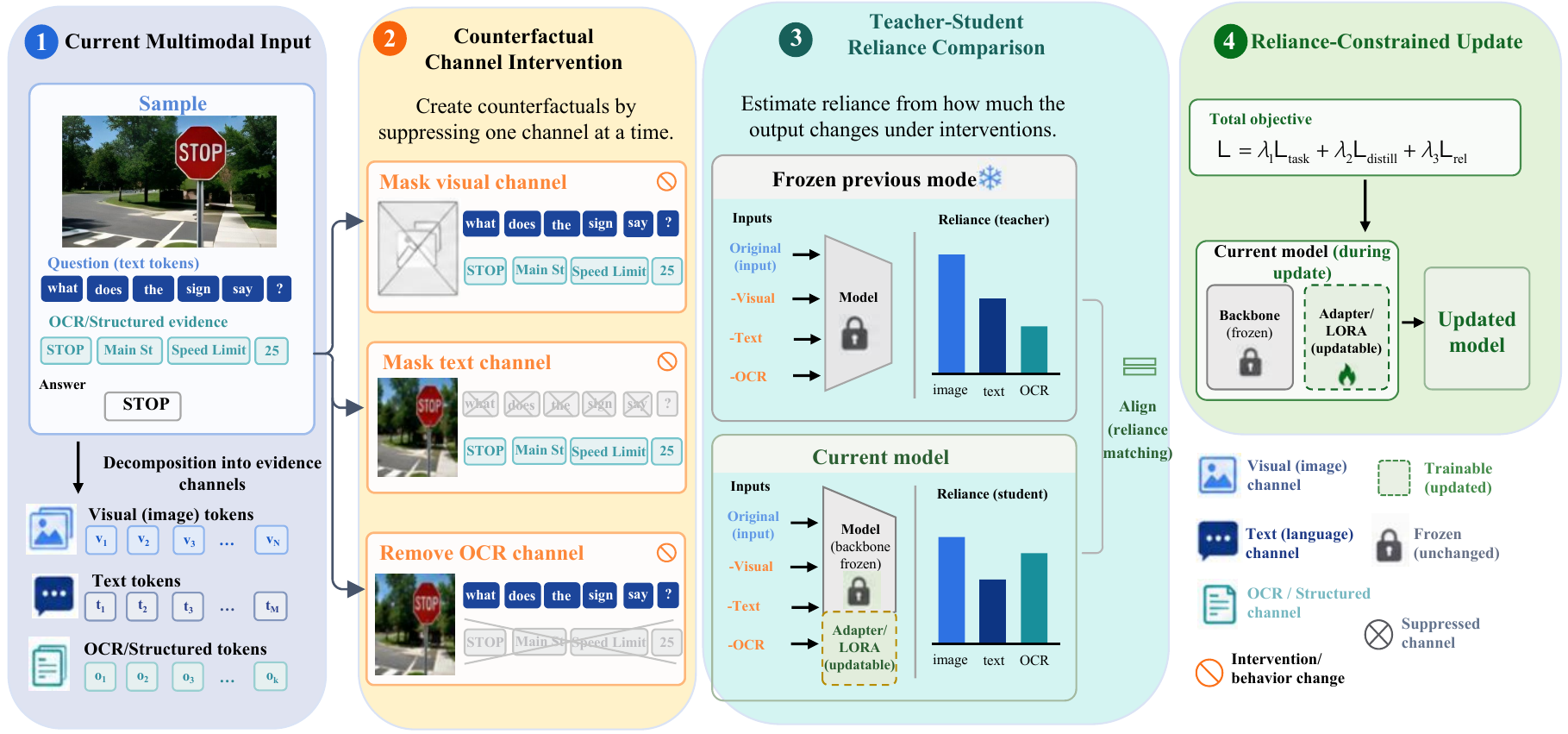}
	\caption{
\textbf{Overview of Reliance-Constrained Continual Learning.}
Given a multimodal input, RCL constructs counterfactual channel interventions, compares teacher--student reliance profiles, and updates the current model with a reliance-preserving objective to mitigate hidden evidence-use forgetting.
}
	\label{fig:rcl_pipeline}
\end{figure}

\subsection{Continual Multimodal Learning Setup}
\label{sec:method_setup}

We consider a sequence of multimodal learning stages
$\{\mathcal D_1,\ldots,\mathcal D_K\}$, where each stage $\mathcal D_k$ contains input-output pairs $(x,y)$. Each input $x$ consists of a set of evidence channels $\mathcal M_x$, such as visual tokens, question tokens, OCR tokens, chart elements, and other task-specific input sources. We write $x=\{x^m:m\in\mathcal M_x\}$, where $x^m$ denotes the component of $x$ associated with channel $m$.

Let $f_\theta$ denote an MLLM with autoregressive answer distribution
$p_\theta(y\mid x)=\prod_{\ell=1}^{L}p_\theta(y_\ell\mid y_{<\ell},x)$.
We use parameter-efficient adaptation and decompose the model parameters as
$\theta=\theta_0\cup\phi$, where $\theta_0$ is the frozen backbone and $\phi$ denotes trainable adapter parameters. At stage $k$, the previous model is frozen as $f_{\theta^-}$ with $\theta^-=\theta_{k-1}$, and the current model $f_\theta$ is optimized on $\mathcal D_k$.

The standard task loss is
\begin{equation}
\label{eq:task_loss}
    \mathcal L_{\rm task}(\theta)
    =
    \mathbb E_{(x,y)\sim\mathcal D_k}
    \left[
    -\sum_{\ell=1}^{L}
    \log p_\theta(y_\ell\mid y_{<\ell},x)
    \right].
\end{equation}
This objective encourages correct answers on the current stage, but it does not constrain which evidence channels the model uses to obtain those answers.

\subsection{Counterfactual Modality Reliance}
\label{sec:method_counterfactual_reliance}

RCL estimates modality reliance through counterfactual evidence suppression. For each channel $m\in\mathcal M_x$, we define an intervention set $\mathcal I_m$. Each intervention $\tau_m\in\mathcal I_m$ suppresses access to channel $m$ while keeping the remaining channels unchanged. For example, an image channel is suppressed by masking visual regions, an OCR channel by removing recognized text tokens, and a chart channel by masking chart elements. The complete intervention construction is given in Appendix~\ref{app:interventions}.

For a model $f_\theta$, we define the counterfactual sensitivity of channel $m$ as
\begin{equation}
\label{eq:counterfactual_score}
\begin{aligned}
    \Delta_{\theta,m}(x,y)
    =
    \frac{1}{|\mathcal I_m|}
    \sum_{\tau_m\in\mathcal I_m}
    \bigg[
    &
    \frac{1}{L}
    \sum_{\ell=1}^{L}
    D_{\rm KL}
    \left(
    p_\theta(\cdot\mid y_{<\ell},x)
    \,\middle\|\,
    p_\theta(\cdot\mid y_{<\ell},\tau_m(x))
    \right)
    \\
    &+
    \beta
    \left(
    \ell_\theta(y\mid \tau_m(x))
    -
    \ell_\theta(y\mid x)
    \right)_+
    \bigg],
\end{aligned}
\end{equation}
where $\ell_\theta(y\mid x)=-\sum_{\ell=1}^{L}\log p_\theta(y_\ell\mid y_{<\ell},x)$, $(z)_+=\max(z,0)$, and $\beta\geq0$ balances output-distribution shift and answer-loss increase. A larger $\Delta_{\theta,m}(x,y)$ indicates that suppressing channel $m$ causes a larger behavioral change, hence the model relies more strongly on that channel for the given sample.

We normalize channel sensitivities into a reliance vector:
\begin{equation}
\label{eq:reliance_vector}
    r_{\theta,m}(x,y)
    =
    \frac{
    \exp\left(\Delta_{\theta,m}(x,y)/\tau_r\right)
    }{
    \sum_{n\in\mathcal M_x}
    \exp\left(\Delta_{\theta,n}(x,y)/\tau_r\right)
    },
    \qquad
    \bm r_\theta(x,y)
    =
    \left[
    r_{\theta,m}(x,y)
    \right]_{m\in\mathcal M_x}.
\end{equation}
Here $\tau_r>0$ is a temperature parameter. The vector $\bm r_\theta(x,y)$ lies in the simplex $\Delta^{|\mathcal M_x|-1}$ and is always computed over the evidence channels available in the current sample. Teacher and student reliance vectors are therefore compared on the same support $\mathcal M_x$.

\subsection{Hidden Forgetting as Reliance Drift}
\label{sec:method_hidden_forgetting}

Accuracy-based evaluation detects whether the final answer changes, but it does not detect whether the evidence path changes. We formalize this discrepancy through modality reliance drift. Given the frozen reference model $f_{\theta^-}$ and the current model $f_\theta$, the sample-level reliance drift is
\begin{equation}
\label{eq:sample_mrd}
    {\rm MRD}(x,y;\theta^-,\theta)
    =
    D_{\rm JS}
    \left(
    \bm r_{\theta^-}(x,y)
    \,\middle\|\,
    \bm r_{\theta}(x,y)
    \right),
\end{equation}
where $D_{\rm JS}$ denotes Jensen--Shannon divergence.

For a previous-stage evaluation set $\mathcal E_j$, hidden forgetting occurs when answer accuracy is preserved while reliance drift is large:
\begin{equation}
\label{eq:hidden_forgetting}
    {\rm HF}_{j\rightarrow k}
    =
    \mathbb I
    \left[
    {\rm Acc}(f_{\theta_j};\mathcal E_j)
    -
    {\rm Acc}(f_{\theta_k};\mathcal E_j)
    \leq \delta_{\rm acc}
    \right]
    \cdot
    \mathbb I
    \left[
    {\rm MRD}_{j\rightarrow k}
    \geq \delta_{\rm rel}
    \right].
\end{equation}
Here ${\rm MRD}_{j\rightarrow k}$ is the average of Eq.~\eqref{eq:sample_mrd} over $\mathcal E_j$, and $\delta_{\rm acc},\delta_{\rm rel}$ are fixed thresholds. This definition separates visible forgetting, measured by answer degradation, from hidden forgetting, measured by evidence-use drift. Additional evaluation metrics are defined in Appendix~\ref{app:metrics}.

\subsection{Reliance-Constrained Continual Update}
\label{sec:method_update}

RCL constrains the current model to preserve the reference model's reliance profile on the current-stage data. For each sample $(x,y)$, the frozen model $f_{\theta^-}$ produces a reference reliance vector $\bm r_{\theta^-}(x,y)$, while the trainable model produces $\bm r_\theta(x,y)$. The reliance preservation loss is
\begin{equation}
\label{eq:reliance_loss}
    \mathcal L_{\rm rel}(\theta)
    =
    \mathbb E_{(x,y)\sim\mathcal D_k}
    \left[
    w_{\theta^-}(x,y)
    D_{\rm JS}
    \left(
    {\rm sg}\!\left[\bm r_{\theta^-}(x,y)\right]
    \,\middle\|\,
    \bm r_{\theta}(x,y)
    \right)
    \right],
\end{equation}
where ${\rm sg}[\cdot]$ denotes stop-gradient. The scalar $w_{\theta^-}(x,y)\in[0,1]$ is a confidence weight computed from the reference model's predictive entropy and intervention sensitivity; its exact form is provided in Appendix~\ref{app:confidence_gate}. This gate prevents unreliable reference predictions from dominating the update.

We also preserve the output behavior of the previous model through token-level prediction distillation:
\begin{equation}
\label{eq:prediction_distillation}
    \mathcal L_{\rm pred}(\theta)
    =
    \mathbb E_{(x,y)\sim\mathcal D_k}
    \left[
    \frac{\tau_{\rm kd}^2}{L}
    \sum_{\ell=1}^{L}
    D_{\rm KL}
    \left(
    p_{\theta^-}^{\tau_{\rm kd}}(\cdot\mid y_{<\ell},x)
    \,\middle\|\,
    p_{\theta}^{\tau_{\rm kd}}(\cdot\mid y_{<\ell},x)
    \right)
    \right],
\end{equation}
where $p_\theta^{\tau_{\rm kd}}$ denotes the softmax distribution over logits divided by the temperature $\tau_{\rm kd}$.

The full objective at stage $k$ is
\begin{equation}
\label{eq:full_objective}
    \min_{\phi}
    \quad
    \mathcal L_k(\phi)
    =
    \mathcal L_{\rm task}
    +
    \lambda_{\rm pred}\mathcal L_{\rm pred}
    +
    \lambda_{\rm rel}\mathcal L_{\rm rel}
    +
    \lambda_{\rm reg}\|\phi-\phi^-\|_2^2 ,
\end{equation}
where $\phi^-$ denotes the adapter parameters from the previous stage. The prediction term preserves what the model answers, while the reliance term preserves how the model uses evidence. RCL does not impose a fixed preference for visual, textual, or OCR evidence. Instead, it preserves the instance-specific reliance profile induced by the previous model. Therefore, when the reference model genuinely relies on visual evidence, RCL discourages drift toward language shortcuts; when the reference model relies primarily on textual evidence, RCL preserves that allocation rather than forcing visual dependence.

\subsection{Training and Inference}
\label{sec:method_training_inference}

At each stage, RCL freezes $f_{\theta^-}$, samples a minibatch from $\mathcal D_k$, constructs counterfactual interventions for the available evidence channels, computes $\bm r_{\theta^-}(x,y)$ and $\bm r_\theta(x,y)$ on the same intervened inputs, and updates only the PEFT parameters $\phi$ using Eq.~\eqref{eq:full_objective}. RCL stores no raw multimodal examples from previous stages. It keeps the previous checkpoint only during the current stage, which is sufficient to provide prediction and reliance targets. After training stage $k$, the updated model becomes the reference for stage $k+1$.

During inference, RCL uses the updated MLLM $f_{\theta_k}$ directly. Counterfactual interventions, frozen-reference forward passes, and reliance estimation are used only during training and evaluation, so the deployed model has the same inference cost as the underlying PEFT-adapted MLLM.

\section{Experiments}
\subsection{Experimental Setup}
\label{sec:experimental_setup}

\begin{table*}[t]
\centering
\small
\setlength{\tabcolsep}{3.0pt}
\renewcommand{\arraystretch}{1.08}
\caption{
\textbf{Main results against common baselines and recent SoTAs.}
Best, second-best, and third-best results are highlighted in red, blue, and green, respectively.
}
\label{tab:main_standard_results}
\resizebox{\linewidth}{!}{
\begin{tabular}{lllccc cc ccc}
\toprule
\multirow{2}{*}{\textbf{Method}} 
& \multirow{2}{*}{\textbf{Backbone}} 
& \multirow{2}{*}{\textbf{Extra signal}} 
& \multicolumn{3}{c}{\textbf{CoIN}} 
& \multicolumn{2}{c}{\textbf{COAST}} 
& \multicolumn{3}{c}{\textbf{MCITlib}} \\
\cmidrule(lr){4-6}\cmidrule(lr){7-8}\cmidrule(lr){9-11}
& & 
& ACC$\uparrow$ & MAA$\uparrow$ & BWT$\uparrow$
& Avg.$\uparrow$ & Fgt.$\downarrow$
& MFN$\uparrow$ & MAA$\uparrow$ & BWT$\uparrow$ \\
\midrule
\rowcolor{groupgray}
\multicolumn{11}{l}{\textbf{Reference settings}} \\
Zero-shot 
& LLaVA-1.5-7B 
& None 
& \NA & 7.12 & \NA 
& \NA & \NA 
& \NA & \NA & \NA \\
Multi-task / Joint 
& LLaVA-1.5-7B 
& All tasks 
& \NA & 57.18 & \NA 
& 52.59 & \NA 
& 58.48 & 63.86 & \NA \\
\midrule
\rowcolor{groupgray}
\multicolumn{11}{l}{\textbf{Standard continual learning and PEFT baselines}} \\
Seq-LoRA 
& LLaVA-1.5-7B 
& None 
& 28.74 & 32.97 & -32.62 
& 30.29 & 20.74 
& 50.22 & 58.09 & -13.41 \\
EWC 
& LLaVA-1.5-7B 
& Fisher 
& 32.90 & 36.93 & -27.46 
& 32.25 & 19.16 
& 49.51 & 56.38 & -12.68 \\
LwF 
& LLaVA-1.5-7B 
& Logit distill. 
& 30.41 & 34.95 & -27.03 
& 32.54 & 20.59 
& 49.10 & 56.25 & -13.05 \\
L2P 
& LLaVA-1.5-7B 
& Prompt 
& 41.86 & 46.72 & -18.49 
& 45.79 & 6.69 
& 50.99 & 57.60 & -8.59 \\
DualPrompt 
& LLaVA-1.5-7B 
& Prompt 
& 42.22 & 47.05 & -18.17 
& 46.27 & 6.61 
& 51.31 & 57.77 & -8.36 \\
CODA-Prompt 
& LLaVA-1.5-7B 
& Prompt 
& 43.50 & 48.26 & -17.29 
& 46.85 & 5.80 
& 52.00 & 58.16 & -7.80 \\
O-LoRA 
& LLaVA-1.5-7B 
& Orth. LoRA 
& 45.83 & 50.26 & -16.58 
& 46.72 & 5.79 
& 52.35 & 58.52 & -9.98 \\
\midrule
\rowcolor{groupgray}
\multicolumn{11}{l}{\textbf{Recent multimodal continual instruction tuning methods}} \\
MoELoRA 
& LLaVA-1.5-7B 
& Expert LoRA 
& 37.13 & 42.76 & -25.91 
& 45.01 & 7.23 
& 49.88 & 57.64 & -10.70 \\
Model Tailor 
& LLaVA-1.5-7B 
& Sparse patch 
& 46.67 & 51.22 & -15.12 
& 45.56 & 5.17 
& 53.05 & 58.98 & -7.48 \\
Continual LLaVA 
& LLaVA-1.5-7B 
& Incr. embed. 
& 48.09 & 52.80 & -13.27 
& 48.73 & 4.54 
& 53.26 & 59.51 & -6.73 \\
CL-MoE 
& LLaVA-1.5-7B 
& MoE 
& 48.90 & 53.48 & -11.92 
& 47.72 & 5.06 
& 49.18 & 57.21 & -12.65 \\
HiDe-LLaVA 
& LLaVA-1.5-7B 
& Decoupling 
& 51.56 & 56.41 & -8.97 
& 48.30 & 4.68 
& 50.18 & 57.53 & -7.58 \\
BranchLoRA 
& LLaVA-1.5-7B 
& Router 
& 44.20 & 49.94 & -20.98 
& 49.11 & 4.56 
& 54.97 & 60.21 & -5.58 \\
ProgLoRA 
& LLaVA-1.5-7B 
& Task recall 
& 59.09 & 62.38 & -6.59 
& 49.79 & 4.11 
& 55.78 & 60.55 & -4.94 \\
ModalPrompt 
& LLaVA-1.5-7B 
& Prompt 
& 50.52 & 55.55 & -10.10 
& 47.95 & 4.80 
& 46.73 & 47.17 & 0.02 \\
SEFE 
& LLaVA-1.5-7B 
& ASD+RegLoRA 
& 53.61 & 58.17 & -8.55 
& 49.36 & 4.31 
& 54.26 & 59.66 & -7.94 \\
MLLM-CL 
& LLaVA-1.5-7B 
& Router 
& 55.51 & 60.28 & -7.17 
& 49.90 & 3.97 
& 55.32 & 60.76 & -4.93 \\
IGVP 
& LLaVA-1.5-7B 
& Projector 
& 55.00 & 59.87 & -7.39 
& 49.45 & 4.23 
& 55.15 & 60.18 & -5.16 \\
\midrule
\rowcolor{groupgray}
\multicolumn{11}{l}{\textbf{Replay or memory-assisted references}} \\
Rehearsal 
& LLaVA-1.5-7B 
& Buffer 
& 42.88 & 47.25 & -18.43 
& 31.58 & 19.52 
& 51.56 & 57.19 & -8.11 \\
DER 
& LLaVA-1.5-7B 
& Buffer+KD 
& 44.35 & 48.76 & -16.76 
& 33.88 & 17.94 
& 52.52 & 57.88 & -7.67 \\
GIFT 
& LLaVA-1.5-7B 
& Synthetic replay 
& 56.26 & 61.51 & -6.89 
& 49.76 & 4.07 
& 56.06 & 60.72 & -4.74 \\
Ask-and-Remember 
& LLaVA-1.5-7B 
& Question replay 
& 57.22 & 62.07 & -6.72 
& 50.02 & 3.90 
& 56.35 & 60.88 & -4.56 \\
\midrule
\rowcolor{groupgray}
\multicolumn{11}{l}{\textbf{Ours}} \\
\textsc{RCL} 
& LLaVA-1.5-7B 
& Reliance 
& \third{65.01} & \third{67.56} & \third{-4.23} 
& \third{52.55} & \third{2.76} 
& \third{58.86} & \third{62.29} & \third{-2.03} \\
\textsc{RCL} 
& LLaVA-1.5-13B 
& Reliance 
& \best{66.87} & \best{69.30} & \best{-3.77} 
& \best{54.22} & \best{2.30} 
& \best{60.18} & \best{63.42} & \best{-1.58} \\
\textsc{RCL} 
& InternVL-Chat-7B 
& Reliance 
& \second{66.08} & \second{68.45} & \second{-3.92} 
& \second{53.50} & \second{2.47} 
& \second{59.62} & \second{62.85} & \second{-1.79} \\
\bottomrule
\end{tabular}
}
\end{table*}

\textbf{Benchmarks and metrics.}
We evaluate \textsc{RCL} on CoIN, COAST, MCITlib, and an evidence-sensitive stream covering VQA, OCR/document, and chart/table reasoning \citep{chen2024coin,cao2024continual,guo2025mcitlib}.
We report official task scores, standard CL metrics, and reliance-aware diagnostics; details are in Appendix~\ref{app:benchmark_and_metrics}.

\textbf{Comparison protocol.}
\textbf{Comparison protocol.}
We use a replay-free, task-id-free protocol: methods store no previous multimodal examples, use no oracle task identity at inference, and follow the same task order and data budget.
We compare against standard CL/PEFT baselines, recent multimodal continual instruction-tuning methods, and separately report replay- or memory-assisted references because they use stored or generated past-task signals.
Detailed method grouping, baseline list, and fairness rules are provided in Appendix~\ref{app:comparison_protocol}.

\textbf{Backbones and implementation.} We use LLaVA-1.5-7B as the default backbone and additionally evaluate LLaVA-1.5-13B and InternVL-Chat-7B to test scale and architecture robustness \citep{liu2024visual,chen2024internvl}. Unless otherwise specified, the vision encoder and language backbone are frozen and only PEFT parameters are updated. All methods are trained with matched data, task order, trainable-parameter budget, and optimization budget whenever possible. For \textsc{RCL}, counterfactual interventions are used only during training and diagnostic evaluation, so the final model has the same inference cost as the underlying PEFT-adapted MLLM. Implementation details, hyperparameters, compute resources, and reproducibility settings are provided in Appendix~\ref{app:implementation_details}.

\subsection{Main Results}
\label{sec:main_results}



\begin{table*}[t]
\centering
\small
\setlength{\tabcolsep}{4.5pt}
\renewcommand{\arraystretch}{1.08}
\caption{
\textbf{Reliance-aware results on the evidence-sensitive stream.}
Avg. is final average answer performance; AF is average forgetting; MRD is modality reliance drift; DEF is dominant evidence flip rate on answer-preserved examples; HFR is hidden forgetting rate; GCR measures grounded-channel retention for visual/OCR/chart/document evidence.
}
\label{tab:main_reliance_results}
\begin{tabular*}{\textwidth}{@{\extracolsep{\fill}}llcccccc@{}}
\toprule
\textbf{Method} 
& \textbf{Backbone} 
& Avg.$\uparrow$ 
& AF$\downarrow$ 
& MRD$\downarrow$ 
& DEF$\downarrow$ 
& HFR$\downarrow$ 
& GCR$\uparrow$ \\
\midrule
EWC 
& LLaVA-1.5-7B 
& 51.96 & 16.06 & 0.257 & 36.7 & 54.0 & 67.4 \\
LwF 
& LLaVA-1.5-7B 
& 50.88 & 16.45 & 0.250 & 36.1 & 52.9 & 67.1 \\
O-LoRA 
& LLaVA-1.5-7B 
& 54.30 & 12.60 & 0.234 & 33.9 & 46.7 & 70.7 \\
MoELoRA 
& LLaVA-1.5-7B 
& 53.62 & 13.22 & 0.223 & 32.6 & 45.4 & 71.3 \\
Continual LLaVA 
& LLaVA-1.5-7B 
& 56.53 & 9.86 & 0.209 & 29.3 & 38.6 & 74.7 \\
HiDe-LLaVA 
& LLaVA-1.5-7B 
& 57.26 & 9.38 & 0.196 & 28.0 & 36.3 & 76.0 \\
BranchLoRA 
& LLaVA-1.5-7B 
& 58.58 & 8.95 & 0.191 & 26.8 & 35.0 & 77.3 \\
ProgLoRA 
& LLaVA-1.5-7B 
& 61.94 & 5.87 & 0.176 & 24.1 & 28.8 & 79.6 \\
ModalPrompt 
& LLaVA-1.5-7B 
& 56.99 & 9.12 & 0.184 & 25.0 & 31.4 & 78.2 \\
SEFE 
& LLaVA-1.5-7B 
& 59.41 & 8.10 & 0.173 & 23.3 & 27.6 & 80.3 \\
MLLM-CL 
& LLaVA-1.5-7B 
& 60.29 & 7.02 & 0.167 & 22.7 & 26.3 & 81.2 \\
GIFT 
& LLaVA-1.5-7B 
& 61.62 & 6.06 & 0.162 & 21.9 & 24.7 & 82.6 \\
Ask-and-Remember 
& LLaVA-1.5-7B 
& 61.90 & 5.80 & 0.159 & 21.6 & 24.0 & 82.9 \\
\midrule
\textsc{RCL} 
& LLaVA-1.5-7B 
& \third{64.06} & \third{4.22} & \third{0.106} & \third{14.8} & \third{12.9} & \third{89.0} \\
\textsc{RCL} 
& LLaVA-1.5-13B 
& \best{65.28} & \best{3.76} & \best{0.093} & \best{13.0} & \best{11.1} & \best{90.6} \\
\textsc{RCL} 
& InternVL-Chat-7B 
& \second{64.75} & \second{3.93} & \second{0.098} & \second{13.7} & \second{11.8} & \second{89.9} \\
\bottomrule
\end{tabular*}
\end{table*}

Table~\ref{tab:main_standard_results} shows that \textsc{RCL} delivers the strongest plasticity--retention trade-off across three backbones, improving final performance while reducing forgetting and keeping BWT closer to zero.
Despite being replay-free, it outperforms replay- or task-recall-based references, highlighting evidence-use preservation as an effective anti-forgetting signal.

Table~\ref{tab:main_reliance_results} shows that strong baselines can preserve answers while still suffering large MRD, DEF, and HFR, revealing substantial evidence-path drift.
\textsc{RCL} consistently reduces reliance drift and dominant-channel flips while improving GCR, confirming that continual MLLM evaluation should assess both answers and their supporting evidence.

\subsection{Ablation Study and Analysis}
\label{sec:ablation}

\paragraph{Single-factor ablation.}
We ablate each major component of \textsc{RCL} on LLaVA-1.5-7B using the same task order, PEFT configuration, training budget, and evaluation protocol as the full model.
As shown in Table~\ref{tab:single_factor_ablation}, removing the reliance loss $\mathcal L_{\rm rel}$ leads to the largest degradation in reliance-aware diagnostics, especially MRD, DEF, and HFR.
This indicates that preserving output behavior or answer accuracy alone is insufficient to maintain stable evidence use during continual adaptation.
Prediction distillation mainly improves conventional retention metrics, while the confidence gate helps filter unreliable teacher reliance targets and evidence-targeted interventions provide more faithful channel-level sensitivity signals.
The weaker performance of random interventions, single-score reliance variants, and visual-only reliance further shows that the gains of \textsc{RCL} do not come from generic perturbation regularization.
Instead, they arise from preserving instance-specific, multi-channel evidence reliance across continual learning stages.

\begin{table*}[t]
\centering
\small
\setlength{\tabcolsep}{3.4pt}
\renewcommand{\arraystretch}{1.08}
\caption{
\textbf{Single-factor ablation on LLaVA-1.5-7B.}
We report standard benchmark performance and evidence-use diagnostics.
Blue numbers indicate degradation relative to the full \textsc{RCL} model.
}
\label{tab:single_factor_ablation}
\resizebox{\linewidth}{!}{
\begin{tabular}{lccc cccccc}
\toprule
\multirow{2}{*}{\textbf{Variant}} 
& \multicolumn{3}{c}{\textbf{Standard continual benchmarks}} 
& \multicolumn{6}{c}{\textbf{Evidence-sensitive stream}} \\
\cmidrule(lr){2-4}\cmidrule(lr){5-10}
& CoIN ACC$\uparrow$ 
& COAST Avg.$\uparrow$ 
& MCIT MFN$\uparrow$
& Avg.$\uparrow$ 
& AF$\downarrow$ 
& MRD$\downarrow$ 
& DEF$\downarrow$ 
& HFR$\downarrow$ 
& GCR$\uparrow$ \\
\midrule
\textsc{RCL} full 
& \best{65.01} 
& \best{52.55} 
& \best{58.86} 
& \best{64.06} 
& \best{4.22} 
& \best{0.106} 
& \best{14.8} 
& \best{12.9} 
& \best{89.0} \\
\midrule
w/o reliance loss $\mathcal L_{\rm rel}$ 
& 62.44\abldelta{(-2.57)} 
& 50.78\abldelta{(-1.77)} 
& 56.91\abldelta{(-1.95)} 
& 61.72\abldelta{(-2.34)} 
& 5.96\abldelta{(+1.74)} 
& 0.184\abldelta{(+0.078)} 
& 25.9\abldelta{(+11.1)} 
& 30.4\abldelta{(+17.5)} 
& 81.7\abldelta{(-7.3)} \\
w/o prediction distillation $\mathcal L_{\rm pred}$ 
& 62.98\abldelta{(-2.03)} 
& 51.10\abldelta{(-1.45)} 
& 57.29\abldelta{(-1.57)} 
& 62.24\abldelta{(-1.82)} 
& 5.31\abldelta{(+1.09)} 
& 0.125\abldelta{(+0.019)} 
& 17.9\abldelta{(+3.1)} 
& 17.2\abldelta{(+4.3)} 
& 86.8\abldelta{(-2.2)} \\
w/o confidence gate $w_{\theta^-}$ 
& 63.28\abldelta{(-1.73)} 
& 51.46\abldelta{(-1.09)} 
& 57.54\abldelta{(-1.32)} 
& 62.83\abldelta{(-1.23)} 
& 5.08\abldelta{(+0.86)} 
& 0.137\abldelta{(+0.031)} 
& 19.4\abldelta{(+4.6)} 
& 20.1\abldelta{(+7.2)} 
& 85.6\abldelta{(-3.4)} \\
w/o entropy factor in gate 
& 63.81\abldelta{(-1.20)} 
& 51.88\abldelta{(-0.67)} 
& 57.92\abldelta{(-0.94)} 
& 63.21\abldelta{(-0.85)} 
& 4.72\abldelta{(+0.50)} 
& 0.126\abldelta{(+0.020)} 
& 17.1\abldelta{(+2.3)} 
& 17.8\abldelta{(+4.9)} 
& 86.9\abldelta{(-2.1)} \\
w/o sensitivity factor in gate 
& 63.66\abldelta{(-1.35)} 
& 51.75\abldelta{(-0.80)} 
& 57.83\abldelta{(-1.03)} 
& 63.08\abldelta{(-0.98)} 
& 4.81\abldelta{(+0.59)} 
& 0.132\abldelta{(+0.026)} 
& 18.6\abldelta{(+3.8)} 
& 18.7\abldelta{(+5.8)} 
& 86.2\abldelta{(-2.8)} \\
random interventions 
& 62.90\abldelta{(-2.11)} 
& 50.96\abldelta{(-1.59)} 
& 57.08\abldelta{(-1.78)} 
& 62.34\abldelta{(-1.72)} 
& 5.42\abldelta{(+1.20)} 
& 0.153\abldelta{(+0.047)} 
& 22.0\abldelta{(+7.2)} 
& 24.8\abldelta{(+11.9)} 
& 83.9\abldelta{(-5.1)} \\
loss-increase only 
& 63.95\abldelta{(-1.06)} 
& 52.05\abldelta{(-0.50)} 
& 58.06\abldelta{(-0.80)} 
& 63.41\abldelta{(-0.65)} 
& 4.63\abldelta{(+0.41)} 
& 0.124\abldelta{(+0.018)} 
& 17.0\abldelta{(+2.2)} 
& 16.3\abldelta{(+3.4)} 
& 86.9\abldelta{(-2.1)} \\
KL-shift only 
& 64.12\abldelta{(-0.89)} 
& 52.12\abldelta{(-0.43)} 
& 58.11\abldelta{(-0.75)} 
& 63.59\abldelta{(-0.47)} 
& 4.54\abldelta{(+0.32)} 
& 0.119\abldelta{(+0.013)} 
& 16.4\abldelta{(+1.6)} 
& 15.7\abldelta{(+2.8)} 
& 87.4\abldelta{(-1.6)} \\
visual-only reliance 
& 63.42\abldelta{(-1.59)} 
& 51.64\abldelta{(-0.91)} 
& 57.68\abldelta{(-1.18)} 
& 62.87\abldelta{(-1.19)} 
& 4.92\abldelta{(+0.70)} 
& 0.141\abldelta{(+0.035)} 
& 20.8\abldelta{(+6.0)} 
& 21.6\abldelta{(+8.7)} 
& 84.7\abldelta{(-4.3)} \\
w/o adapter regularization 
& 64.31\abldelta{(-0.70)} 
& 52.19\abldelta{(-0.36)} 
& 58.27\abldelta{(-0.59)} 
& 63.55\abldelta{(-0.51)} 
& 4.53\abldelta{(+0.31)} 
& 0.118\abldelta{(+0.012)} 
& 16.2\abldelta{(+1.4)} 
& 15.1\abldelta{(+2.2)} 
& 87.8\abldelta{(-1.2)} \\
\bottomrule
\end{tabular}
}
\end{table*}

\paragraph{Can Accuracy Preservation Hide Evidence-Use Forgetting?}
\label{sec:answer_drop_mrd_bubble}

\begin{figure}[H]
    \centering
    \includegraphics[width=0.72\linewidth]{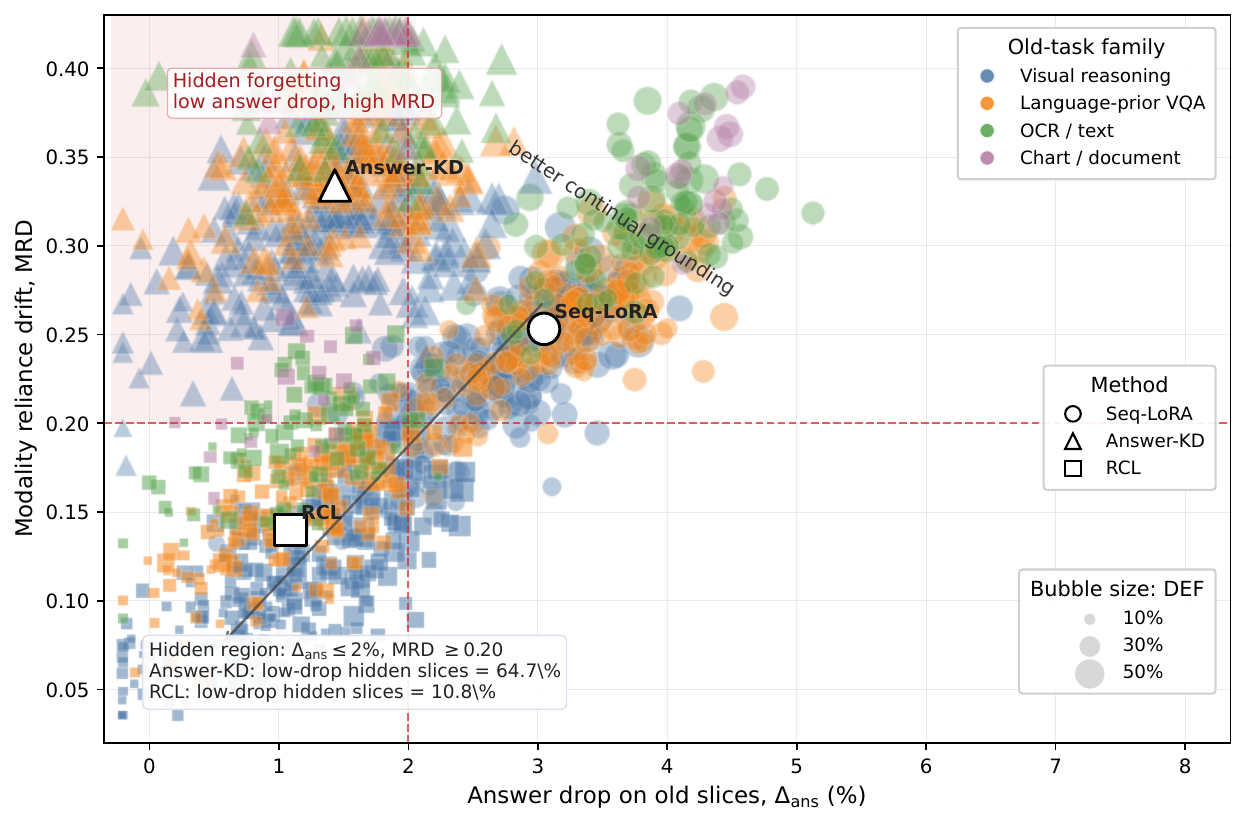}
    \caption{
    \textbf{Answer-level retention can hide evidence-use forgetting.}
    Protocol details are in \hyperref[app:detail_answer_drop_mrd]{Appendix~\ref{app:experimental_details}, \emph{Answer-drop--MRD bubble protocol}}.
    }
    \label{fig:answer_drop_mrd_bubble}
\end{figure}

Answer-level retention may conceal whether an old task is still supported by the same multimodal evidence.
We analyze all old-to-later transitions in the evidence-sensitive stream and compare clean answer drop, modality reliance drift (MRD), and dominant evidence flip rate (DEF) on answer-preserved examples.
Figure~\ref{fig:answer_drop_mrd_bubble} shows that \textsc{Answer-KD} moves many points leftward by reducing the average answer drop from $3.41\%$ to $1.82\%$, but still leaves $64.7\%$ of low-drop slices in the hidden-forgetting region with an average DEF of $35.9\%$.
By contrast, \textsc{RCL} reduces MRD from $0.238$ to $0.108$, lowers DEF from $35.9\%$ to $14.6\%$, and decreases hidden forgetting from $64.7\%$ to $10.8\%$.
The largest reductions occur on OCR-, chart-, and document-centric slices, indicating that reliance preservation is most valuable when the correct answer depends on structured evidence rather than language priors.

\paragraph{Is Counterfactual Reliance Estimation Worth the Training Cost?}
\label{sec:cost_performance_pareto}

\begin{figure}[H]
    \centering
    \includegraphics[width=0.78\linewidth]{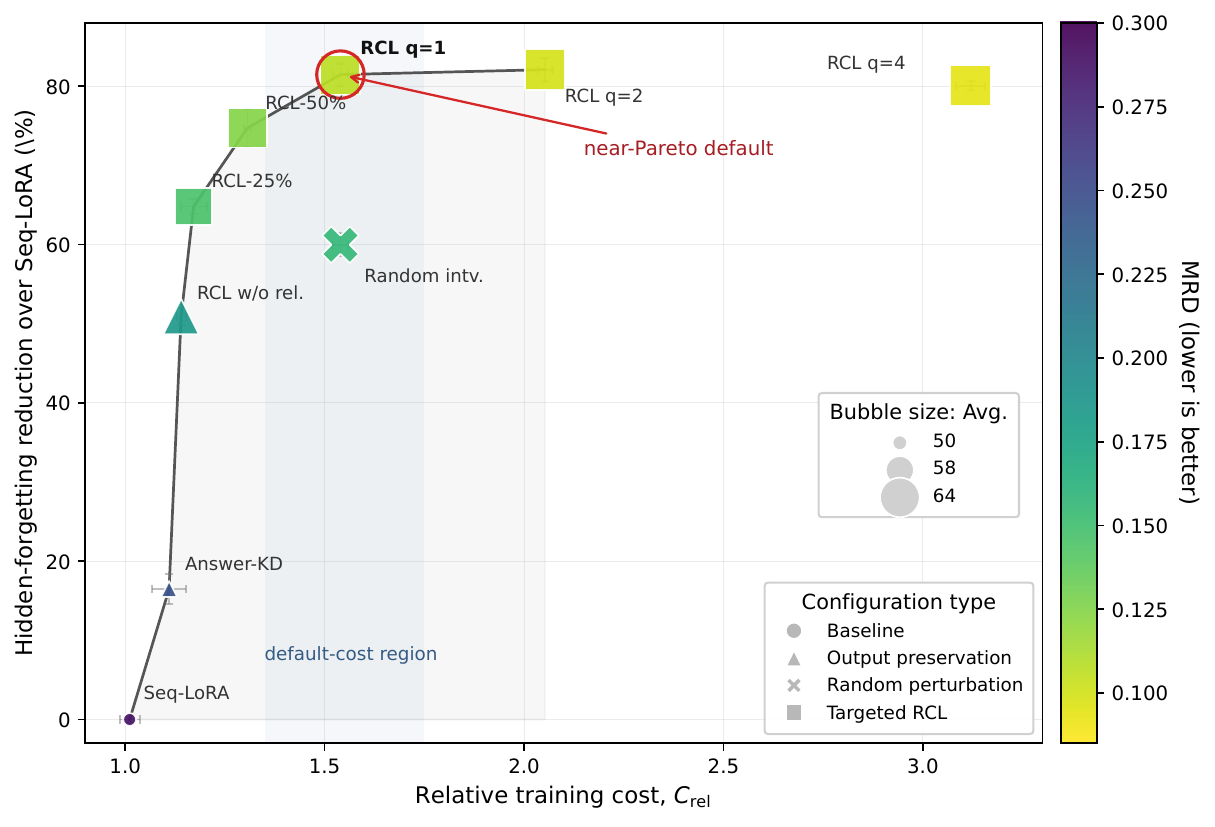}
    \caption{
    \textbf{Cost--performance Pareto analysis.}
    The $x$-axis is relative training cost, the $y$-axis is HFR reduction over Seq-LoRA, bubble size is final average accuracy, and color indicates MRD.
    The default \textsc{RCL} setting lies near the Pareto frontier.
    Protocol details are in \hyperref[app:detail_cost_performance_pareto]{Appendix~\ref{app:experimental_details}, \emph{Cost--performance Pareto protocol}}.
    }
    \label{fig:cost_performance_pareto}
\end{figure}

Because counterfactual reliance estimation introduces extra training-time forward passes, we test whether its hidden-forgetting gains justify the added cost.
We compare sequential tuning, output distillation, random intervention regularization, sparse \textsc{RCL}, and \textsc{RCL} variants with different intervention budgets using relative GPU cost and HFR reduction.
Figure~\ref{fig:cost_performance_pareto} shows that output distillation adds little cost but reduces HFR by only $13.9\%$, confirming that output preservation alone is insufficient.
Random interventions incur similar cost to \textsc{RCL} but achieve only $59.7\%$ HFR reduction and retain higher MRD, indicating that targeted evidence suppression is necessary.
The default setting $q_{\rm train}=1$ reduces HFR by $79.0\%$ at $1.56\times$ training cost, while increasing to $q_{\rm train}=2$ and $q_{\rm train}=4$ yields only marginal gains at substantially higher cost.
Thus, one targeted intervention per available evidence channel already provides a strong reliance signal with no inference-time overhead.

\section{Conclusion}

This paper identifies hidden evidence-use forgetting in continual multimodal learning, where correct answers may be retained while grounding behavior silently drifts.
We propose \textsc{RCL}, a replay-free framework that preserves evidence-reliance profiles through counterfactual channel interventions without adding inference-time cost.
Experiments show that answer-level retention alone can mask substantial reliance drift, while \textsc{RCL} improves both performance retention and grounding stability.
These results suggest that continual MLLMs should be evaluated by not only what they answer, but also how they use multimodal evidence.
Future work may extend reliance-aware adaptation to longer streams and richer evidence structures.


\bibliographystyle{plainnat}
\bibliography{reference}

\newpage

\appendix

\section{Additional Method Details}
\label{app:method_details}

\subsection{Counterfactual Intervention Construction}
\label{app:interventions}

RCL estimates reliance through evidence-suppressing interventions. Each intervention targets one evidence channel while keeping the remaining channels unchanged. We instantiate the intervention set $\mathcal I_m$ according to the channel type.

\paragraph{Visual channel.}
For image-based inputs, $\mathcal I_{\rm img}$ contains region masking and visual-token removal. Region masking replaces selected image patches with a neutral value. Visual-token removal removes the corresponding visual embeddings before multimodal fusion. For datasets with object annotations, regions are sampled from annotated objects. For datasets without object annotations, regions are sampled from uniformly partitioned image patches.

\paragraph{Question-text channel.}
For question or instruction text, $\mathcal I_{\rm txt}$ contains content-word masking. Stop words and formatting tokens are preserved. This intervention suppresses semantic cues from the question while keeping the prompt format stable.

\paragraph{OCR channel.}
For OCR-based inputs, $\mathcal I_{\rm ocr}$ contains OCR-token removal and OCR-token masking. OCR-token removal deletes recognized text tokens from the input sequence. OCR-token masking replaces recognized text tokens with a neutral placeholder token.

\paragraph{Chart and table channels.}
For chart and table reasoning, $\mathcal I_{\rm chart}$ contains axis-label masking, legend masking, cell-value masking, and plotted-element masking. These interventions suppress numerical and structural evidence while preserving the global input format.

\paragraph{Video channel.}
For video inputs, $\mathcal I_{\rm vid}$ contains frame masking and sparse frame removal. Frame masking suppresses visual content in selected frames. Sparse frame removal drops selected frame embeddings before multimodal fusion.

For each minibatch, RCL samples a fixed number of interventions from $\mathcal I_m$ for every available channel $m\in\mathcal M_x$. The same intervened inputs are shared by the frozen reference model and the current model, ensuring that the reliance loss compares teacher and student behavior under identical counterfactual conditions.

\subsection{Confidence Gate for Reliance Targets}
\label{app:confidence_gate}

The reliance loss in Eq.~\eqref{eq:reliance_loss} uses a confidence weight $w_{\theta^-}(x,y)$ to reduce the influence of unreliable reference targets. The gate combines predictive confidence and intervention sensitivity.

The reference predictive entropy is
\begin{equation}
\label{eq:app_entropy}
    H_{\theta^-}(x,y)
    =
    \frac{1}{L}
    \sum_{\ell=1}^{L}
    H\!\left(
    p_{\theta^-}(\cdot\mid y_{<\ell},x)
    \right),
\end{equation}
where $H(\cdot)$ denotes Shannon entropy. The total reference intervention sensitivity is
\begin{equation}
\label{eq:app_total_sensitivity}
    A_{\theta^-}(x,y)
    =
    \sum_{m\in\mathcal M_x}
    \Delta_{\theta^-,m}(x,y).
\end{equation}
The confidence weight is defined as
\begin{equation}
\label{eq:app_confidence_weight}
    w_{\theta^-}(x,y)
    =
    \sigma
    \left(
    \frac{\tau_H-H_{\theta^-}(x,y)}{\gamma_H}
    \right)
    \cdot
    \sigma
    \left(
    \frac{A_{\theta^-}(x,y)-\tau_A}{\gamma_A}
    \right),
\end{equation}
where $\sigma(\cdot)$ is the sigmoid function, $\tau_H$ and $\tau_A$ are fixed thresholds, and $\gamma_H,\gamma_A>0$ are temperature parameters. The first factor assigns larger weight to low-entropy reference predictions. The second factor assigns larger weight to samples where the reference model exhibits measurable dependence on at least one evidence channel.

\subsection{Diagnostic Attribution Analysis}
\label{app:diagnostic_attribution}

RCL does not use gradient-based attribution in the training objective. We use attribution only as an evaluation-time diagnostic to verify whether reliance preservation is consistent with local sensitivity in the model representation.

Let $h_\theta^m(x)$ denote the hidden representation associated with channel $m$. The attribution score for channel $m$ is
\begin{equation}
\label{eq:app_grad_attr}
    G_{\theta,m}(x,y)
    =
    \frac{
    \left\|
    h_\theta^m(x)
    \odot
    \nabla_{h_\theta^m(x)}
    \ell_\theta(y\mid x)
    \right\|_1
    }{
    {\rm dim}\!\left(h_\theta^m(x)\right)
    }.
\end{equation}
Since $G_{\theta,m}$ is never differentiated through during training, RCL avoids second-order optimization. We report attribution agreement by comparing the normalized attribution vector
$\bm g_\theta(x,y)$ with the counterfactual reliance vector $\bm r_\theta(x,y)$ using rank correlation and Jensen--Shannon divergence.

\subsection{Task-Preserving Perturbation Analysis}
\label{app:task_preserving_perturbations}

In addition to evidence-suppressing interventions, we use task-preserving perturbations for robustness analysis. These perturbations are not part of the main reliance loss. They evaluate whether a model is overly sensitive to changes that should not alter the answer.

For each channel $m$, let $\mathcal P_m$ be a set of task-preserving perturbations. Examples include mild image noise, small resolution changes, harmless crop-and-resize transformations, question paraphrases that preserve meaning, and OCR formatting changes that preserve recognized text. The invariance sensitivity is
\begin{equation}
\label{eq:app_invariance_sensitivity}
    B_{\theta,m}(x,y)
    =
    \frac{1}{|\mathcal P_m|}
    \sum_{\pi_m\in\mathcal P_m}
    \frac{1}{L}
    \sum_{\ell=1}^{L}
    D_{\rm KL}
    \left(
    p_\theta(\cdot\mid y_{<\ell},x)
    \,\middle\|\,
    p_\theta(\cdot\mid y_{<\ell},\pi_m(x))
    \right).
\end{equation}
A grounded and stable model should have high sensitivity to evidence-suppressing interventions and low sensitivity to task-preserving perturbations.

\subsection{Efficient Training}
\label{app:efficient_training}

The computational overhead of RCL comes from counterfactual forward passes through the frozen reference model and the current model. We control this overhead with three design choices.

First, for each minibatch, RCL samples a small fixed number of interventions per channel. Second, the same counterfactual inputs are used for both teacher and student models. Third, reliance estimation is applied only to the channels present in the current sample. Therefore, if a sample contains $|\mathcal M_x|$ channels and $q$ interventions are sampled per channel, the reliance module uses $q|\mathcal M_x|$ counterfactual views for that sample.

The teacher model is evaluated under stop-gradient. The student model receives gradients only through the reliance vector $\bm r_\theta(x,y)$ computed from counterfactual output shifts. No gradient-based attribution term is used in the training objective.

\subsection{Training Procedure}
\label{app:training_procedure}

At continual stage $k$, RCL performs the following procedure.

\begin{enumerate}
    \item Freeze the previous model $f_{\theta^-}=f_{\theta_{k-1}}$.
    \item Initialize the current trainable model with $\theta=\theta^-$ and update only PEFT parameters $\phi$.
    \item Sample a minibatch $\mathcal B_k$ from the current-stage dataset $\mathcal D_k$.
    \item For each $(x,y)\in\mathcal B_k$ and each channel $m\in\mathcal M_x$, construct counterfactual inputs $\tau_m(x)$.
    \item Compute reference reliance $\bm r_{\theta^-}(x,y)$ and current reliance $\bm r_\theta(x,y)$ using Eq.~\eqref{eq:counterfactual_score} and Eq.~\eqref{eq:reliance_vector}.
    \item Compute $\mathcal L_{\rm task}$, $\mathcal L_{\rm pred}$, and $\mathcal L_{\rm rel}$ using Eq.~\eqref{eq:task_loss}, Eq.~\eqref{eq:prediction_distillation}, and Eq.~\eqref{eq:reliance_loss}.
    \item Update $\phi$ by minimizing Eq.~\eqref{eq:full_objective}.
    \item After stage $k$ finishes, set $\theta_k=\theta$ and use $f_{\theta_k}$ as the frozen reference for stage $k+1$.
\end{enumerate}

\subsection{Evaluation Metrics}
\label{app:metrics}

We evaluate continual multimodal learning from two perspectives: answer retention and reliance preservation.

\paragraph{Final average accuracy.}
After the final stage $K$, the final average accuracy is
\begin{equation}
\label{eq:app_average_accuracy}
    {\rm AA}_K
    =
    \frac{1}{K}
    \sum_{j=1}^{K}
    {\rm Acc}(f_{\theta_K};\mathcal E_j),
\end{equation}
where $\mathcal E_j$ is the evaluation set for stage $j$.

\paragraph{Average forgetting.}
The standard average forgetting score is
\begin{equation}
\label{eq:app_average_forgetting}
    {\rm AF}_K
    =
    \frac{1}{K-1}
    \sum_{j=1}^{K-1}
    \left[
    \max_{s\in\{j,\ldots,K\}}
    {\rm Acc}(f_{\theta_s};\mathcal E_j)
    -
    {\rm Acc}(f_{\theta_K};\mathcal E_j)
    \right].
\end{equation}

\paragraph{Reliance drift.}
The final reliance drift is
\begin{equation}
\label{eq:app_final_mrd}
    {\rm MRD}_K
    =
    \frac{1}{K-1}
    \sum_{j=1}^{K-1}
    \frac{1}{|\mathcal E_j|}
    \sum_{(x,y)\in\mathcal E_j}
    D_{\rm JS}
    \left(
    \bm r_{\theta_j}(x,y)
    \,\middle\|\,
    \bm r_{\theta_K}(x,y)
    \right).
\end{equation}

\paragraph{Channel-specific drift.}
For channel $m$, the channel-specific drift from stage $j$ to the final model is
\begin{equation}
\label{eq:app_channel_drift}
    \Delta_{j\rightarrow K}^{m}
    =
    \frac{1}{|\mathcal E_j^m|}
    \sum_{(x,y)\in\mathcal E_j^m}
    \left[
    r_{\theta_K,m}(x,y)
    -
    r_{\theta_j,m}(x,y)
    \right],
\end{equation}
where $\mathcal E_j^m=\{(x,y)\in\mathcal E_j:m\in\mathcal M_x\}$. A negative value for the visual channel indicates a drift away from visual evidence.

\paragraph{Hidden forgetting rate.}
The hidden forgetting rate is
\begin{equation}
\label{eq:app_hfr}
    {\rm HFR}_K
    =
    \frac{1}{K-1}
    \sum_{j=1}^{K-1}
    {\rm HF}_{j\rightarrow K},
\end{equation}
where ${\rm HF}_{j\rightarrow K}$ is defined in Eq.~\eqref{eq:hidden_forgetting}. A reliable continual MLLM should achieve high final average accuracy, low average forgetting, low reliance drift, and low hidden forgetting rate.

\section{Additional Experimental Details}
\label{app:exp_details}

\subsection{Details of the Preliminary Study}
\label{app:prelim_details}

This subsection provides the experimental details omitted from Section~\ref{sec:preliminary}. 
The goal of the preliminary study is diagnostic: it exposes a failure mode of standard replay-free multimodal continual adaptation, rather than evaluating the proposed method.

\paragraph{Continual stream.}
We construct an eight-stage multimodal learning stream:
ScienceQA, GQA, VQAv2, OK-VQA, TextVQA, OCR-VQA, ChartQA, and DocVQA.
These stages cover visual reasoning, compositional VQA, knowledge-intensive VQA, OCR-based question answering, chart understanding, and document understanding.
After learning each stage $j$, we save the resulting checkpoint $f_j$.
After learning a later stage $k>j$, we evaluate the current checkpoint $f_k$ on the previous-stage evaluation set $\mathcal E_j$.
No previous training examples are replayed during adaptation.

\paragraph{Baselines.}
We use two simple baselines to avoid introducing any component of our method into the exploratory study.
The first baseline, \textbf{Seq-LoRA}, updates only parameter-efficient adaptation parameters on the current-stage task loss:
\begin{equation}
\label{eq:app_seq_lora}
    \mathcal L_{\rm Seq}
    =
    \mathbb E_{(x,y)\sim\mathcal D_k}
    \left[
    -\log p_{\theta}(y\mid x)
    \right].
\end{equation}
The second baseline, \textbf{Answer-KD}, additionally preserves the previous checkpoint's output distribution on the current-stage data:
\begin{equation}
\label{eq:app_answer_kd}
    \mathcal L_{\rm KD}
    =
    \mathcal L_{\rm Seq}
    +
    \lambda_{\rm kd}
    \mathbb E_{(x,y)\sim\mathcal D_k}
    \left[
    D_{\rm KL}
    \left(
    p_{\theta^-}^{\tau}(\cdot\mid x)
    \,\middle\|\,
    p_{\theta}^{\tau}(\cdot\mid x)
    \right)
    \right],
\end{equation}
where $\theta^-$ denotes the previous checkpoint and $\tau$ is the distillation temperature.
This baseline preserves output behavior but imposes no constraint on which evidence channel supports the prediction.

\paragraph{Post-hoc channel ablations.}
For each example $(x,y)$, we define a set of available evidence channels $\mathcal C_x$.
Depending on the task, this set may include visual regions, OCR tokens, document text, chart elements, and question-side language priors.
For each channel $m\in\mathcal C_x$, we construct an ablated input $\tau_m(x)$ that suppresses channel $m$ while keeping other channels unchanged.
For example, visual ablation masks image tokens or visual regions; OCR/text ablation removes recognized text or document tokens; chart ablation masks chart elements; and question-prior ablation removes or neutralizes evidence-bearing lexical cues in the question.
These ablations are used only after training for diagnosis.
They are not used as training objectives, constraints, supervision signals, or data augmentation.

For a checkpoint $f$, we define the diagnostic channel sensitivity as
\begin{equation}
\label{eq:app_prelim_sensitivity}
    s_m(f;x,y)
    =
    \left[
    \ell_f(y\mid \tau_m(x))
    -
    \ell_f(y\mid x)
    \right]_+,
\end{equation}
where $\ell_f(y\mid x)$ is the negative log-likelihood of the ground-truth answer and $[z]_+=\max(z,0)$.
The normalized evidence-use vector is
\begin{equation}
\label{eq:app_prelim_q}
    q_m(f;x,y)
    =
    \frac{
    s_m(f;x,y)
    }{
    \sum_{n\in\mathcal C_x}s_n(f;x,y)+\epsilon
    },
    \qquad
    \bm q(f;x,y)
    =
    \left[
    q_m(f;x,y)
    \right]_{m\in\mathcal C_x}.
\end{equation}
The small constant $\epsilon$ prevents numerical instability when all channel sensitivities are close to zero.
The vector $\bm q(f;x,y)$ is only a post-hoc diagnostic proxy for evidence use.

\paragraph{Metrics.}
For each old-to-new transition $(j,k)$ with $k>j$, we report three quantities.
The clean accuracy drop $A_{j\rightarrow k}$ measures conventional answer forgetting.
The evidence-sensitivity drift $G_{j\rightarrow k}$ measures how much the normalized evidence-use vector changes between checkpoints.
The dominant-evidence flip rate $H_{j\rightarrow k}$ is computed only on examples for which both $f_j$ and $f_k$ predict the correct answer.
Thus, $H_{j\rightarrow k}$ isolates cases where the answer is preserved but the dominant evidence channel changes.
In Figure~\ref{fig:preliminary_hidden_forgetting}, we mark a transition as falling into the hidden-drift region when
\begin{equation}
\label{eq:app_hidden_drift_region}
    A_{j\rightarrow k} \leq 2\%,
    \qquad
    G_{j\rightarrow k} \geq 0.20.
\end{equation}
These thresholds are used only to visualize the qualitative phenomenon and are not used by our method.

\paragraph{Evaluation slices and aggregation.}
To avoid hiding heterogeneous behavior under task-level averages, each previous-stage evaluation set is partitioned into several diagnostic slices, including easy, medium, hard, rare, compositional, and format-sensitive subsets.
Each point in Figure~\ref{fig:preliminary_hidden_forgetting}(a) corresponds to one tuple of old task, later task, random seed, evaluation slice, and training baseline.
The x-axis is $A_{j\rightarrow k}$, the y-axis is $G_{j\rightarrow k}$, the marker size is $H_{j\rightarrow k}$, the marker shape denotes the baseline, and the color denotes the old-task family.
Figure~\ref{fig:preliminary_hidden_forgetting}(b) focuses on Answer-KD.
Each cell corresponds to evaluating an old task after a later stage is learned; color indicates the dominant-evidence flip rate, and bubble size indicates the clean accuracy drop.

\paragraph{Why this does not leak the proposed method.}
The preliminary study does not optimize reliance preservation, intervention consistency, evidence-channel matching, or any loss introduced in the main method.
The baselines are trained using only task supervision and, for Answer-KD, output-level distillation.
Channel ablations are applied strictly after training to diagnose the learned checkpoints.
Therefore, the study serves only as motivation: it reveals that output preservation can mask evidence-use drift, which motivates the reliance-preservation objective introduced in Section~\ref{sec:method}.

\paragraph{Relation-level alignment utility protocol.}
\label{app:detail_obs_volcano}
For the observability volcano plot in Figure~\ref{fig:obs_volcano}, each point corresponds to one modality--semantic relation \(r=(m,n,s)\).
We compute the final observability score \(O_r=O_{m,n,s}\) after training and evaluate the relation-conditioned validation performance of three strategies: Task-only FedAvg, Uniform Alignment, and FedSO.
The relation-level gains are defined as
\begin{equation}
G^{\mathrm{UA}}_{r}
=
\operatorname{Score}^{\mathrm{UA}}_{r}
-
\operatorname{Score}^{\mathrm{Task}}_{r},
\qquad
G^{\mathrm{FedSO}}_{r}
=
\operatorname{Score}^{\mathrm{FedSO}}_{r}
-
\operatorname{Score}^{\mathrm{Task}}_{r}.
\label{eq:obs_volcano_gain}
\end{equation}
A positive value indicates beneficial transfer, while a negative value indicates negative transfer.
All relation-level quantities are averaged over three random seeds before plotting.
The horizontal axis shows \(O_r\), the vertical axis shows the corresponding gain, point size indicates relation support, point color encodes observability, and red markers highlight negative-transfer relations.
For summary statistics, we define low-observability relations as \(O_r<0.35\) and high-observability relations as \(O_r>0.65\).

\paragraph{Low-observability negative-transfer protocol.}
\label{app:detail_low_obs_violin}
For Figure~\ref{fig:low_obs_violin}, we group all modality--semantic relations according to their final observability scores:
\begin{equation}
\text{Low-O}: O_r < 0.35,
\qquad
\text{Mid-O}: 0.35 \le O_r < 0.65,
\qquad
\text{High-O}: O_r \ge 0.65.
\label{eq:obs_group_def}
\end{equation}
For each group and each method \(\mathcal A\), we report the relation-conditioned gain over Task-only FedAvg:
\begin{equation}
G_r^{(\mathcal A)}
=
\operatorname{Score}^{(\mathcal A)}_r
-
\operatorname{Score}^{\mathrm{Task}}_r,
\qquad
\mathcal A \in
\{\mathrm{Task},\mathrm{UA},\mathrm{FedSO}\}.
\label{eq:violin_relation_gain}
\end{equation}
The Task-only baseline is therefore centered at zero and serves as the reference distribution.
Each violin aggregates all modality--semantic relations from the corresponding observability group across benchmark datasets and seeds.
We overlay the group mean and interquartile range on each violin.
The negative-transfer rate is computed as
\begin{equation}
\operatorname{NTR}^{(\mathcal A)}_{\mathcal G}
=
\frac{1}{|\mathcal G|}
\sum_{r\in\mathcal G}
\mathbf{1}\!\left[
G_r^{(\mathcal A)}<0
\right],
\label{eq:negative_transfer_rate}
\end{equation}
where \(\mathcal G\) denotes one of the three observability groups.

\paragraph{Relation-geometry visualization protocol.}
\label{app:detail_relation_geometry}
For Figure~\ref{fig:relation_geometry_umap}, we visualize the geometry of learned modality--semantic prototypes.
For each modality \(m\) and semantic unit \(s\), we extract the final global prototype
\begin{equation}
\bar{\mu}_{m,s}
=
\frac{
\sum_{i}
e_{i,m,s} n_{i,m,s}\rho_{i,m,s}\tilde{\mu}_{i,m,s}
}{
\sum_{i}
e_{i,m,s} n_{i,m,s}\rho_{i,m,s}
+\varepsilon
},
\label{eq:appendix_global_proto}
\end{equation}
where \(e_{i,m,s}\), \(n_{i,m,s}\), \(\rho_{i,m,s}\), and \(\tilde{\mu}_{i,m,s}\) follow the semantic sketch definitions in the main method section.
We collect all prototypes \(\{\bar{\mu}_{m,s}\}_{m,s}\) and project them into two dimensions using UMAP with cosine distance.
The same projection procedure and visualization set are used for Uniform Alignment and FedSO.
Each point in the plot corresponds to one prototype \(\bar{\mu}_{m,s}\), point color indicates the semantic unit, and marker shape indicates the modality.
We draw an edge between \(\bar{\mu}_{m,s}\) and \(\bar{\mu}_{n,s}\) when the corresponding relation is highly observable, i.e.,
\begin{equation}
O_{m,n,s}>\tau_h.
\label{eq:appendix_high_obs_edge}
\end{equation}
To quantify the geometry, we report the average cross-modal prototype distance over high-observability relations:
\begin{equation}
D_{\mathrm{high}}
=
\frac{
1
}{
|\mathcal R_{\mathrm{high}}|
}
\sum_{(m,n,s)\in\mathcal R_{\mathrm{high}}}
\left\|
\bar{\mu}_{m,s}
-
\bar{\mu}_{n,s}
\right\|_2,
\qquad
\mathcal R_{\mathrm{high}}
=
\{(m,n,s): O_{m,n,s}>\tau_h\}.
\label{eq:high_obs_proto_distance}
\end{equation}
We also compute an inter-semantic margin:
\begin{equation}
\operatorname{Margin}
=
\frac{1}{S}
\sum_{s=1}^{S}
\min_{s'\ne s}
\left\|
\bar{\mu}_{s}
-
\bar{\mu}_{s'}
\right\|_2,
\qquad
\bar{\mu}_{s}
=
\frac{1}{|\mathcal M_s|}
\sum_{m\in\mathcal M_s}
\bar{\mu}_{m,s},
\label{eq:inter_semantic_margin}
\end{equation}
where \(\mathcal M_s\) is the set of modalities that provide a valid prototype for semantic unit \(s\).
A smaller \(D_{\mathrm{high}}\) indicates tighter alignment among reliable relations, while a larger margin indicates better separation between semantic units.

\subsection{Detailed Experimental Setup}
\label{app:benchmark_and_metrics}

\subsubsection{Benchmark Suites}
\label{app:benchmark_suites}

We evaluate \textsc{RCL} on four benchmark suites that together cover standard multimodal continual learning, recent multimodal continual instruction tuning, and the evidence-use drift targeted by our method.

\paragraph{CoIN.}
CoIN \citep{chen2024coin} is used as a primary continual instruction tuning benchmark for MLLMs. 
It contains diverse multimodal tasks, including visual question answering, OCR reasoning, object recognition, image classification, and referring expression comprehension. 
Following the official protocol, we train the model sequentially over the task stream and evaluate all seen tasks after each stage. 
This benchmark is important because many recent MCIT methods report CoIN results, making it a suitable setting for direct SOTA comparison.

\paragraph{COAST.}
We use the COAST benchmark from Continual LLaVA \citep{cao2024continual} to evaluate three complementary continual adaptation scenarios: domain-incremental learning, capability-incremental learning, and dataset-incremental learning. 
The domain-incremental setting evaluates robustness across visual domains such as natural images, documents, charts, medical-style images, and scientific visual content. 
The capability-incremental setting evaluates whether the model can sequentially acquire different multimodal abilities, including recognition, OCR, chart reasoning, visual conversation, and complex reasoning. 
The dataset-incremental setting evaluates whether the model can adapt to new datasets without catastrophically forgetting previously learned ones.

\paragraph{MCITlib.}
We further evaluate on MCITlib \citep{guo2025mcitlib}, which provides leakage-aware and reproducible multimodal continual instruction tuning protocols. 
We use its UCIT, MLLM-DCL, and MLLM-ACL settings. 
UCIT evaluates unified continual instruction tuning across heterogeneous task families. 
MLLM-DCL emphasizes domain continual learning, where the task format may remain similar but the visual or semantic domain changes. 
MLLM-ACL emphasizes ability continual learning, where the model must sequentially acquire different reasoning and perception abilities. 
This suite is used to ensure that our conclusions are not specific to a single benchmark design.

\paragraph{Evidence-sensitive stream.}
To directly evaluate hidden evidence-use forgetting, we construct an eight-stage stream:
\[
\text{ScienceQA} \rightarrow \text{GQA} \rightarrow \text{VQAv2} \rightarrow \text{OK-VQA} \rightarrow 
\text{TextVQA} \rightarrow \text{OCR-VQA} \rightarrow \text{ChartQA} \rightarrow \text{DocVQA}.
\]
This stream is designed to contain tasks with different dominant evidence sources. 
ScienceQA and GQA emphasize visual reasoning and compositional grounding; VQAv2 and OK-VQA include strong language-prior and external-knowledge effects; TextVQA and OCR-VQA require OCR grounding; ChartQA requires chart-element and numerical reasoning; DocVQA requires document text and layout understanding. 
This makes the stream suitable for testing whether a continual MLLM preserves the evidence path behind retained answers.

\begin{table*}[t]
\centering
\small
\setlength{\tabcolsep}{4pt}
\caption{
\textbf{Benchmark overview.}
The main paper reports compact benchmark names, while this appendix provides task composition and evaluation purpose.
}
\label{tab:benchmark_overview}
\begin{tabular}{p{0.16\linewidth}p{0.28\linewidth}p{0.34\linewidth}p{0.14\linewidth}}
\toprule
\textbf{Suite} & \textbf{Continual setting} & \textbf{Representative tasks} & \textbf{Purpose} \\
\midrule
CoIN 
& Continual instruction tuning 
& VQA, OCR-QA, recognition, classification, referring expression comprehension 
& Main MCIT comparison \\
COAST 
& Domain-, capability-, and dataset-incremental adaptation 
& ChartQA, DocVQA, IconQA, medical QA, conversation, reasoning, REC, VQA 
& Cross-setting robustness \\
MCITlib 
& Leakage-aware MCIT protocol 
& UCIT, MLLM-DCL, MLLM-ACL 
& Reproducible benchmark comparison \\
Evidence-sensitive stream 
& Reliance-aware continual adaptation 
& ScienceQA, GQA, VQAv2, OK-VQA, TextVQA, OCR-VQA, ChartQA, DocVQA 
& Hidden evidence-use forgetting \\
\bottomrule
\end{tabular}
\end{table*}

\subsubsection{Dataset Preprocessing and Answer Normalization}
\label{app:data_preprocessing}

For all benchmarks, we follow the official train/validation/test splits whenever available. 
If a benchmark provides multiple task orders, we report the average over the official orders. 
For datasets without an official continual split, we use the original training set for the corresponding continual stage and reserve the official validation set for evaluation. 
To prevent dataset size from dominating the sequential dynamics, we cap each stage at the same maximum number of training examples unless the official benchmark protocol specifies otherwise.

For closed-form VQA tasks, we apply standard VQA answer normalization, including lowercasing, punctuation removal, article removal, and number normalization. 
For OCR- and document-centric tasks, we additionally normalize whitespace, line breaks, and common OCR artifacts. 
For chart and table QA, we normalize numerical answers by allowing minor formatting differences such as commas, percentage signs, and decimal-place variants. 
For open-ended generative tasks, we use the official evaluator of the corresponding benchmark. 
When GPT-assisted scoring is required by the benchmark, we use it only for evaluation and never for training or hyperparameter selection.

\subsubsection{Task-Level Metrics}
\label{app:task_metrics}

We report the official metric for each task family:
\begin{itemize}
    \item \textbf{VQA-style tasks:} exact-match or soft VQA accuracy, depending on the dataset evaluator.
    \item \textbf{Classification and recognition:} top-1 accuracy.
    \item \textbf{Referring expression comprehension:} $\mathrm{Acc@0.5}$ based on box IoU.
    \item \textbf{OCR and document QA:} normalized exact match and ANLS when supported by the official evaluator.
    \item \textbf{Chart and table QA:} relaxed exact match for numerical answers and official ChartQA-style accuracy.
    \item \textbf{Open-ended instruction following:} official GPT-assisted or rule-based score, following the benchmark protocol.
\end{itemize}

When a benchmark reports a single aggregated score, we use the official aggregation. 
When a benchmark reports separate task-family scores, we report both the task-family score and the macro-average across task families.

\subsubsection{Continual Learning Metrics}
\label{app:cl_metrics}

Let $a_{i,j}$ denote the score on evaluation set $\mathcal E_i$ after training stage $j$. 
For a $K$-stage stream, we report immediate task acquisition, final retained performance, average forgetting, backward transfer, and mean accuracy over the learning trajectory:
\begin{equation}
\label{eq:app_standard_cl_metrics}
\begin{aligned}
\mathrm{MFT}
&=
\frac{1}{K}\sum_{i=1}^{K} a_{i,i},
\\
\mathrm{MFN}
&=
\frac{1}{K}\sum_{i=1}^{K} a_{i,K},
\\
\mathrm{AF}
&=
\frac{1}{K-1}\sum_{i=1}^{K-1}
\left(
\max_{j\in\{i,\ldots,K\}}a_{i,j}
-
a_{i,K}
\right),
\\
\mathrm{BWT}
&=
\frac{1}{K-1}\sum_{i=1}^{K-1}
\left(a_{i,K}-a_{i,i}\right),
\\
\mathrm{MAA}
&=
\frac{1}{K}
\sum_{j=1}^{K}
\frac{1}{j}
\sum_{i=1}^{j}a_{i,j}.
\end{aligned}
\end{equation}
Here MFT measures how well the model learns each new stage when it is first trained; MFN measures final retained performance; AF measures the amount of forgetting; BWT measures whether later training improves or harms previous tasks; and MAA measures average performance throughout the whole continual trajectory.

\subsubsection{Reliance-Aware Metrics}
\label{app:reliance_metrics}

Standard continual learning metrics only evaluate whether the final answer is correct. 
Since our work studies hidden evidence-use forgetting, we additionally evaluate how the model's reliance profile changes during continual adaptation.

For each example $(x,y)$ and checkpoint $\theta_j$, we compute a normalized reliance vector $\bm r_{\theta_j}(x,y)$ over available evidence channels. 
The evidence channels may include image tokens, question tokens, OCR tokens, document text, chart elements, table cells, layout regions, and language-prior cues. 
The channel set is sample-dependent and is denoted by $\mathcal M_x$.

\paragraph{Modality reliance drift.}
The reliance drift from stage $i$ to the final checkpoint $K$ is
\begin{equation}
\label{eq:app_metric_mrd}
\mathrm{MRD}_{i\rightarrow K}
=
\frac{1}{|\mathcal E_i|}
\sum_{(x,y)\in \mathcal E_i}
D_{\rm JS}
\left(
\bm r_{\theta_i}(x,y)
\middle\|
\bm r_{\theta_K}(x,y)
\right).
\end{equation}
We report the stream-level score
\begin{equation}
\label{eq:app_metric_mrd_avg}
\mathrm{MRD}_{K}
=
\frac{1}{K-1}
\sum_{i=1}^{K-1}
\mathrm{MRD}_{i\rightarrow K}.
\end{equation}
Lower MRD indicates that the final model preserves the evidence-use pattern of earlier checkpoints.

\paragraph{Dominant evidence flip rate.}
The dominant evidence flip rate isolates examples where the answer remains correct but the dominant evidence channel changes:
\begin{equation}
\label{eq:app_metric_def}
\mathrm{DEF}_{i\rightarrow K}
=
\mathbb E_{(x,y)\in\mathcal E_i}
\left[
\mathbf{1}
\left[
\arg\max_{m\in\mathcal M_x} r_{\theta_i,m}(x,y)
\neq
\arg\max_{m\in\mathcal M_x} r_{\theta_K,m}(x,y)
\right]
\middle|
f_{\theta_i}(x)=f_{\theta_K}(x)=y
\right].
\end{equation}
A high DEF means that answer preservation is partly misleading: the model still answers correctly but uses a different evidence source.

\paragraph{Hidden forgetting rate.}
We define hidden forgetting as the event where answer degradation is small but reliance drift is large:
\begin{equation}
\label{eq:app_metric_hfr}
\mathrm{HFR}_{K}
=
\frac{1}{K-1}
\sum_{i=1}^{K-1}
\mathbf{1}
\left[
a_{i,i}-a_{i,K}\leq \delta_{\rm acc}
\right]
\cdot
\mathbf{1}
\left[
\mathrm{MRD}_{i\rightarrow K}\geq \delta_{\rm rel}
\right].
\end{equation}
Unless otherwise specified, we set $\delta_{\rm acc}=2\%$ and $\delta_{\rm rel}=0.20$ for all experiments. 
These thresholds are fixed before evaluation and are not tuned per dataset.

\paragraph{Channel-specific drift.}
For a channel $m$, we report
\begin{equation}
\label{eq:app_metric_channel_drift}
\Delta^{m}_{i\rightarrow K}
=
\frac{1}{|\mathcal E_i^m|}
\sum_{(x,y)\in \mathcal E_i^m}
\left[
r_{\theta_K,m}(x,y)-r_{\theta_i,m}(x,y)
\right],
\end{equation}
where $\mathcal E_i^m=\{(x,y)\in\mathcal E_i:m\in\mathcal M_x\}$. 
Negative visual, OCR, chart, or document drift indicates that the final model relies less on the corresponding grounded channel than the earlier checkpoint.

\subsubsection{General Multimodal Ability Evaluation}
\label{app:general_eval}

After the final continual stage, we additionally evaluate the final checkpoint on general multimodal ability benchmarks, including POPE, MME, MMBench, and SEED-Bench, following MCITlib \citep{guo2025mcitlib}. 
These benchmarks are not used for training or hyperparameter selection. 
They serve as post-hoc checks that reliance preservation does not merely overfit the continual stream while degrading hallucination robustness, perception, commonsense reasoning, or general visual instruction following.

\subsection{Comparison Protocol}
\label{app:comparison_protocol}

\subsubsection{Method Groups}
\label{app:method_groups}

We group compared methods according to the information and memory they require.

\paragraph{Replay-free and task-id-free methods.}
This is the primary comparison group. 
All methods in this group are trained sequentially without storing previous examples and are evaluated without oracle task identity. 
This group includes Seq-LoRA, EWC, LwF, L2P, DualPrompt, CODA-Prompt, O-LoRA, MoELoRA, Model Tailor, Continual LLaVA, CL-MoE, HiDe-LLaVA, BranchLoRA, ProgLoRA, ModalPrompt, SEFE, MLLM-CL, IGVP, and \textsc{RCL}.
This table is the main table for fair comparison.

\paragraph{Replay- or memory-assisted methods.}
Some methods store old examples, generate replay data, or preserve past-task questions. 
We report them separately because they use additional past-task signals unavailable to strictly replay-free methods. 
This group includes rehearsal, DER \citep{buzzega2020dark}, GIFT \citep{wu2025gift}, and Ask-and-Remember \citep{marouf2025ask}. 
For rehearsal and DER, we use a $1\%$ memory buffer unless otherwise specified by the official protocol. 
For synthetic replay methods, we follow the official generation budget and report the additional generated-sample count.

\paragraph{Oracle or task-aware methods.}
If a method requires oracle task identifiers, a task-specific expert selector, or separate task-specific heads at inference time, we report it in an oracle table. 
Oracle results are useful upper references but are not mixed with the main replay-free task-id-free comparison. 
When the official implementation provides both oracle and non-oracle variants, we report both.

\subsubsection{Fairness Rules}
\label{app:fairness_rules}

We apply the following fairness rules across all directly comparable methods:
\begin{itemize}
    \item \textbf{Same task order.} All methods are trained on the same continual stream and evaluated after the same stages.
    \item \textbf{Same data budget.} Each method sees the same current-stage training examples. Replay-free methods do not access previous-stage examples.
    \item \textbf{Same backbone.} Direct comparisons are made under the same MLLM backbone.
    \item \textbf{Matched trainable parameters.} When a method allows rank, prompt length, or expert-count adjustment, we tune it so that the trainable parameter count is within $5\%$ of \textsc{RCL}.
    \item \textbf{Matched optimization budget.} Methods are trained with the same number of epochs, batch size, and total update steps whenever their official protocol permits.
    \item \textbf{No hidden task identity.} The main table does not provide oracle task IDs at inference time.
    \item \textbf{Separate memory accounting.} Methods using stored or generated replay are reported with their memory or generation budget.
    \item \textbf{Same evaluation interventions.} Reliance-aware diagnostics use the same post-hoc counterfactual intervention set for every method.
\end{itemize}

\subsubsection{Baseline Implementation Notes}
\label{app:baseline_notes}

\paragraph{Seq-LoRA.}
Seq-LoRA updates LoRA adapters sequentially on each stage using only the current-stage task loss. 
It is the closest minimal baseline to \textsc{RCL} because it uses the same backbone and adapter placement but does not preserve predictions or reliance profiles.

\paragraph{Answer-KD / LwF.}
Answer-KD preserves the previous checkpoint's output distribution on current-stage samples using token-level KL divergence. 
It tests whether output preservation alone is sufficient to prevent hidden evidence-use forgetting.

\paragraph{Prompt-based continual learning.}
L2P, DualPrompt, and CODA-Prompt are adapted to the multimodal instruction tuning setting by inserting learnable prompts into the language-side input or adapter-compatible prompt slots. 
When task identifiers are not available, prompt selection uses the official key-query matching mechanism.

\paragraph{Adapter- and LoRA-based MCIT methods.}
O-LoRA, BranchLoRA, ProgLoRA, Model Tailor, HiDe-LLaVA, ModalPrompt, CL-MoE, SEFE, and MLLM-CL are implemented following their official designs whenever code is available. 
If official code is unavailable, we reproduce the method from the paper using the same backbone and training schedule used for other baselines.

\paragraph{Replay-based methods.}
For rehearsal and DER, we maintain a fixed-size buffer selected uniformly from previous stages. 
For GIFT and Ask-and-Remember, we follow the original synthetic-data or question-only replay protocol and report the replay budget separately.

\subsection{Implementation Details}
\label{app:implementation_details}

\subsubsection{Backbones}
\label{app:backbones}

We use three MLLM backbones:
\begin{itemize}
    \item \textbf{LLaVA-1.5-7B} \citep{liu2024visual}: the default backbone for main experiments because it is widely used in recent MCIT work.
    \item \textbf{LLaVA-1.5-13B} \citep{liu2024visual}: used to evaluate whether reliance preservation remains effective at a larger model scale.
    \item \textbf{InternVL-Chat-7B} \citep{chen2024internvl}: used to test whether \textsc{RCL} generalizes beyond the LLaVA model family.
\end{itemize}
Unless otherwise specified, the visual encoder, language backbone, and multimodal projector are frozen. 
Only PEFT parameters are updated during continual adaptation.

\subsubsection{PEFT Configuration}
\label{app:peft_config}

For \textsc{RCL}, we use LoRA adapters on attention projection layers. 
The default configuration is:
\[
r=16,\qquad
\alpha=32,\qquad
p_{\rm dropout}=0.05.
\]
For larger backbones, we keep the same LoRA rank unless otherwise specified. 
For baselines with tunable parameter budgets, such as prompt length, expert number, or adapter rank, we adjust their configuration to match the trainable parameter count of \textsc{RCL} within $5\%$.

\subsubsection{Optimization}
\label{app:optimization}

We train all methods using AdamW with bfloat16 mixed precision. 
The default optimization configuration is:
\[
\text{global batch size}=32,\quad
\text{warm-up ratio}=3\%,\quad
\text{gradient clipping}=1.0,\quad
\text{weight decay}=0.
\]
For LLaVA-style backbones, the default learning rate is $2\times 10^{-4}$ for LoRA parameters. 
For InternVL-style backbones, we use $1\times 10^{-4}$ for stability. 
For COAST and MCITlib settings with official hyperparameters, we additionally report the official-setting result when it differs from the matched-budget setting. 
Each continual stage is trained for two epochs unless the official benchmark protocol requires a fixed update budget.

\subsubsection{\textsc{RCL} Hyperparameters}
\label{app:rcl_hyperparams}

The default \textsc{RCL} objective is
\[
\mathcal L
=
\mathcal L_{\rm task}
+
\lambda_{\rm pred}\mathcal L_{\rm pred}
+
\lambda_{\rm rel}\mathcal L_{\rm rel}
+
\lambda_{\rm reg}\|\phi-\phi^-\|_2^2.
\]
We use the following default values:
\[
\tau_r=0.5,\quad
\tau_{\rm kd}=2.0,\quad
\beta=1.0,\quad
\lambda_{\rm pred}=0.5,\quad
\lambda_{\rm rel}=1.0,\quad
\lambda_{\rm reg}=10^{-4}.
\]
We select these values using the validation split of the first stream only and keep them fixed for all later stages and all benchmark suites. 
This avoids using previous-task validation sets as a hidden replay signal.

\subsubsection{Counterfactual Intervention Budget}
\label{app:intervention_budget}

For each sample, \textsc{RCL} identifies the available evidence channels $\mathcal M_x$. 
During training, we sample one intervention per available channel:
\[
q_{\rm train}=1.
\]
During diagnostic evaluation, we use four interventions per available channel:
\[
q_{\rm eval}=4.
\]
The same intervened inputs are used for the previous checkpoint and current checkpoint, so teacher--student reliance vectors are compared under identical counterfactual views.

\subsubsection{Evidence Channel Taxonomy}
\label{app:evidence_channels}

We instantiate evidence channels according to the input structure of each dataset:
\begin{itemize}
    \item \textbf{Image channel:} image patches, visual tokens, or annotated regions.
    \item \textbf{Question channel:} instruction tokens and content-bearing question tokens.
    \item \textbf{OCR channel:} recognized text tokens and OCR bounding boxes.
    \item \textbf{Document channel:} document text spans, line blocks, and layout regions.
    \item \textbf{Chart/table channel:} axes, legends, plotted marks, table cells, and numerical entries.
    \item \textbf{Language-prior channel:} lexical cues in the question that can induce shortcut answers.
\end{itemize}
For datasets without explicit OCR, document, or chart annotations, the corresponding channels are omitted rather than approximated. 
Reliance vectors are always normalized over the channels available in the current sample.

\subsubsection{Counterfactual Intervention Construction}
\label{app:counterfactual_intervention_construction}

We use evidence-suppressing interventions rather than task-preserving augmentations for reliance estimation. 
The goal is to measure whether suppressing a channel changes the model's output distribution or answer likelihood.

\paragraph{Image interventions.}
We mask image patches, remove selected visual tokens, or replace visual regions with neutral image statistics. 
If object annotations are available, regions are sampled from annotated objects; otherwise, patches are sampled from a uniform grid.

\paragraph{OCR interventions.}
We remove OCR tokens, replace them with a neutral placeholder, or mask OCR spans while preserving the prompt template. 
When OCR bounding boxes are available, visual regions corresponding to OCR spans are also masked in an aligned intervention.

\paragraph{Document interventions.}
We mask document lines, paragraph blocks, or layout regions. 
For DocVQA-style inputs, we preserve page order and prompt structure while suppressing the selected document evidence.

\paragraph{Chart and table interventions.}
We mask axis labels, legends, plotted elements, cell values, and numerical entries. 
For tables, we distinguish header masking from body-cell masking because these two perturbations often affect different reasoning paths.

\paragraph{Question-prior interventions.}
We mask content words in the question while preserving stop words, formatting tokens, and instruction templates. 
This intervention estimates whether the model relies on question-side language priors rather than grounded visual or textual evidence.

\subsubsection{Compute Resources}
\label{app:compute_resources}

All main experiments are run on $8$ NVIDIA A100-80GB GPUs. 
For LLaVA-1.5-7B experiments, each continual stage uses approximately $6$--$10$ GPU hours depending on the dataset size and intervention budget. 
For LLaVA-1.5-13B, each stage uses approximately $12$--$18$ GPU hours. 
For InternVL-Chat-7B, each stage uses approximately $8$--$12$ GPU hours. 
The full main benchmark suite requires approximately $1{,}200$--$1{,}600$ A100 GPU hours, including three random seeds, ablations, and diagnostic evaluation. 
Reliance diagnostics are more expensive than standard answer evaluation because they require counterfactual forward passes, but they are used only during training and evaluation, not at deployment time.

\subsubsection{Random Seeds and Reporting}
\label{app:seeds_reporting}

Unless otherwise specified, we report the mean and standard deviation over three random seeds. 
The seeds affect model initialization of PEFT parameters, task sampling order when multiple official orders are available, data shuffling, and intervention sampling. 
For main tables, we report mean performance. 
For ablations and diagnostic plots, we include standard deviation or confidence intervals when space permits. 
All reliance-aware metrics are computed on the same evaluation examples across methods to avoid intervention-sampling artifacts.

\subsubsection{Inference Cost}
\label{app:inference_cost}

\textsc{RCL} uses counterfactual interventions and the frozen previous checkpoint only during training and diagnostic evaluation. 
After each continual stage, the deployed model is the updated PEFT-adapted MLLM. 
No reference model, intervention module, or reliance estimator is used at inference time. 
Therefore, \textsc{RCL} has the same inference-time architecture, memory footprint, and latency as the underlying LoRA-adapted backbone.

\subsubsection{Hyperparameter Sensitivity}
\label{app:hyperparam_sensitivity}

We evaluate sensitivity to three key hyperparameters:
\[
\lambda_{\rm rel}\in\{0.25,0.5,1.0,2.0\},\qquad
\tau_r\in\{0.25,0.5,1.0\},\qquad
q_{\rm train}\in\{1,2,4\}.
\]
We report both answer-level metrics and reliance-aware metrics. 
This analysis tests whether reliance preservation is robust to the strength of the reliance loss, the sharpness of the reliance distribution, and the number of counterfactual interventions used during training.

\subsubsection{Ablation Settings}
\label{app:ablation_settings}

We evaluate the following ablations:
\begin{itemize}
    \item \textbf{w/o reliance loss:} removes $\mathcal L_{\rm rel}$ and keeps only task learning and prediction distillation.
    \item \textbf{w/o prediction distillation:} removes $\mathcal L_{\rm pred}$ and keeps task learning and reliance preservation.
    \item \textbf{w/o confidence gate:} sets $w_{\theta^-}(x,y)=1$ for all examples.
    \item \textbf{visual-only reliance:} preserves reliance only for visual channels.
    \item \textbf{text/OCR-only reliance:} preserves reliance only for text and OCR channels.
    \item \textbf{random interventions:} replaces evidence-targeted interventions with random token or patch masking.
    \item \textbf{larger intervention budget:} increases $q_{\rm train}$ from $1$ to $2$ or $4$.
\end{itemize}
The ablations are designed to test whether the performance gain comes from reliance preservation itself, from output distillation, from the confidence gate, or merely from generic perturbation regularization.

\subsection{Main Experiment Details}
\label{app:experimental_details}

\phantomsection
\paragraph{Answer-drop--MRD bubble protocol.}
\label{app:detail_answer_drop_mrd}
This paragraph provides the full protocol for Figure~\ref{fig:answer_drop_mrd_bubble}.
For every method $\mathcal A$, random seed, old stage $j$, later checkpoint $k>j$, and diagnostic slice $\mathcal E_{j,s}$, we compute the clean answer drop
\begin{equation}
\label{eq:app_answer_drop}
\Delta_{\mathrm{ans}}^{(\mathcal A)}(j,k,s)
=
\mathrm{Acc}\!\left(f_{\theta_j}^{(\mathcal A)};\mathcal E_{j,s}\right)
-
\mathrm{Acc}\!\left(f_{\theta_k}^{(\mathcal A)};\mathcal E_{j,s}\right).
\end{equation}
We compute modality reliance drift using the reliance vectors defined in Section~\ref{sec:method_counterfactual_reliance}:
\begin{equation}
\label{eq:app_mrd_transition}
\mathrm{MRD}^{(\mathcal A)}_{j\rightarrow k,s}
=
\frac{1}{|\mathcal E_{j,s}|}
\sum_{(x,y)\in\mathcal E_{j,s}}
D_{\rm JS}
\left(
\bm r_{\theta_j}^{(\mathcal A)}(x,y)
\middle\|
\bm r_{\theta_k}^{(\mathcal A)}(x,y)
\right).
\end{equation}
The dominant evidence flip rate is computed only on examples whose answers are preserved:
\begin{equation}
\label{eq:app_def_transition}
\mathrm{DEF}^{(\mathcal A)}_{j\rightarrow k,s}
=
\mathbb E_{(x,y)\in\mathcal E_{j,s}}
\left[
\mathbf{1}
\left[
\arg\max_{m\in\mathcal M_x} r_{\theta_j,m}^{(\mathcal A)}(x,y)
\neq
\arg\max_{m\in\mathcal M_x} r_{\theta_k,m}^{(\mathcal A)}(x,y)
\right]
\middle|
f_{\theta_j}^{(\mathcal A)}(x)=f_{\theta_k}^{(\mathcal A)}(x)=y
\right].
\end{equation}
Each bubble in Figure~\ref{fig:answer_drop_mrd_bubble} corresponds to one tuple
\[
(\mathcal A,\; \mathrm{seed},\; j,\; k,\; s).
\]
The horizontal coordinate is Eq.~\eqref{eq:app_answer_drop}, the vertical coordinate is Eq.~\eqref{eq:app_mrd_transition}, and the bubble area is proportional to Eq.~\eqref{eq:app_def_transition}.
We mark a slice as hidden forgetting when
\begin{equation}
\label{eq:app_hidden_region_threshold}
\Delta_{\mathrm{ans}}^{(\mathcal A)}(j,k,s)\leq 2\%,
\qquad
\mathrm{MRD}^{(\mathcal A)}_{j\rightarrow k,s}\geq 0.20.
\end{equation}
These thresholds are fixed before evaluation and are used only to visualize the discrepancy between answer preservation and evidence-use preservation.
All reported summary statistics are averaged over old stages, later checkpoints, diagnostic slices, and three random seeds.

\phantomsection
\paragraph{Dominant-channel Sankey protocol.}
\label{app:detail_dominant_channel_sankey}
This paragraph provides the construction details for Figure~\ref{fig:dominant_channel_sankey}.
For each method $\mathcal A$ and old stage $j$, we retain only answer-preserved examples:
\begin{equation}
\label{eq:app_answer_preserved_set}
\Omega^{(\mathcal A)}_j
=
\left\{
(x,y)\in\mathcal E_j:
f_{\theta_j}^{(\mathcal A)}(x)=f_{\theta_K}^{(\mathcal A)}(x)=y
\right\},
\end{equation}
where $K$ denotes the final continual stage.
For each example, we identify the dominant evidence channel before and after continual adaptation:
\begin{equation}
\label{eq:app_dominant_channel}
c_j^{(\mathcal A)}(x,y)
=
\arg\max_{m\in\mathcal M_x}
r_{\theta_j,m}^{(\mathcal A)}(x,y),
\qquad
c_K^{(\mathcal A)}(x,y)
=
\arg\max_{m\in\mathcal M_x}
r_{\theta_K,m}^{(\mathcal A)}(x,y).
\end{equation}
The Sankey transition probability from old dominant channel $u$ to final dominant channel $v$ is
\begin{equation}
\label{eq:app_sankey_transition}
T_{u\rightarrow v}^{(\mathcal A)}
=
\frac{
\sum_{j=1}^{K-1}
\sum_{(x,y)\in\Omega^{(\mathcal A)}_j}
\mathbf{1}\!\left[c_j^{(\mathcal A)}(x,y)=u\right]
\mathbf{1}\!\left[c_K^{(\mathcal A)}(x,y)=v\right]
}{
\sum_{j=1}^{K-1}
\sum_{(x,y)\in\Omega^{(\mathcal A)}_j}
\mathbf{1}\!\left[c_j^{(\mathcal A)}(x,y)=u\right]
+\epsilon
}.
\end{equation}
Diagonal retention is computed as
\begin{equation}
\label{eq:app_diagonal_retention}
\mathrm{DiagRet}^{(\mathcal A)}
=
\frac{
\sum_{j=1}^{K-1}
\sum_{(x,y)\in\Omega^{(\mathcal A)}_j}
\mathbf{1}\!\left[
c_j^{(\mathcal A)}(x,y)=c_K^{(\mathcal A)}(x,y)
\right]
}{
\sum_{j=1}^{K-1}
|\Omega^{(\mathcal A)}_j|
+\epsilon
}.
\end{equation}
Shortcut-oriented flip rate into the question-prior channel $q$ is
\begin{equation}
\label{eq:app_question_prior_flip}
\mathrm{QFlip}^{(\mathcal A)}
=
\frac{
\sum_{j=1}^{K-1}
\sum_{(x,y)\in\Omega^{(\mathcal A)}_j}
\mathbf{1}\!\left[
c_j^{(\mathcal A)}(x,y)\neq q
\right]
\mathbf{1}\!\left[
c_K^{(\mathcal A)}(x,y)=q
\right]
}{
\sum_{j=1}^{K-1}
|\Omega^{(\mathcal A)}_j|
+\epsilon
}.
\end{equation}
In Figure~\ref{fig:dominant_channel_sankey}, ribbon width is proportional to Eq.~\eqref{eq:app_sankey_transition}.
Red-tinted ribbons correspond to transitions whose final channel is the question-prior channel.
We additionally report channel-specific diagonal retention for OCR, chart, and document channels to quantify whether structured evidence remains the dominant support for answer-preserved examples.

\phantomsection
\paragraph{Cost--performance Pareto protocol.}
\label{app:detail_cost_performance_pareto}
This paragraph provides the full protocol for Figure~\ref{fig:cost_performance_pareto}.
We compare the following configurations: Seq-LoRA, \textsc{Answer-KD}, random intervention regularization, sparse \textsc{RCL}, and \textsc{RCL} with intervention budgets $q_{\rm train}\in\{1,2,4\}$.
For each configuration $\mathcal A$, the relative training cost is
\begin{equation}
\label{eq:app_relative_training_cost}
C_{\mathrm{rel}}(\mathcal A)
=
\frac{
\mathrm{GPUHours}(\mathcal A)
}{
\mathrm{GPUHours}(\mathrm{Seq\mbox{-}LoRA})
}.
\end{equation}
The hidden-forgetting reduction over Seq-LoRA is
\begin{equation}
\label{eq:app_hfr_reduction}
\mathrm{HFR\mbox{-}Red.}(\mathcal A)
=
\frac{
\mathrm{HFR}(\mathrm{Seq\mbox{-}LoRA})
-
\mathrm{HFR}(\mathcal A)
}{
\mathrm{HFR}(\mathrm{Seq\mbox{-}LoRA})
}
\times 100.
\end{equation}
A method is Pareto-dominated if another method achieves both lower or equal relative cost and higher or equal HFR reduction, with at least one strict improvement:
\begin{equation}
\label{eq:app_pareto_dominance}
\mathcal A \prec \mathcal B
\quad\Longleftrightarrow\quad
C_{\mathrm{rel}}(\mathcal B)\leq C_{\mathrm{rel}}(\mathcal A),
\quad
\mathrm{HFR\mbox{-}Red.}(\mathcal B)\geq \mathrm{HFR\mbox{-}Red.}(\mathcal A),
\end{equation}
with one inequality strict.
In Figure~\ref{fig:cost_performance_pareto}, the $x$-axis is Eq.~\eqref{eq:app_relative_training_cost}, the $y$-axis is Eq.~\eqref{eq:app_hfr_reduction}, bubble area is proportional to final average answer performance, and color encodes MRD.
The default \textsc{RCL} configuration uses $q_{\rm train}=1$, i.e., one targeted intervention per available evidence channel.
Because counterfactual interventions are used only during training and diagnostic evaluation, all points in the figure have the same inference-time architecture and latency as their corresponding PEFT-adapted backbone.

\section{Additional Experimental Results}
\label{app:additional_results}

\subsection{Does RCL Prevent Dominant Evidence-Channel Flips?}

\begin{figure*}[t]
    \centering
    \includegraphics[width=0.96\linewidth]{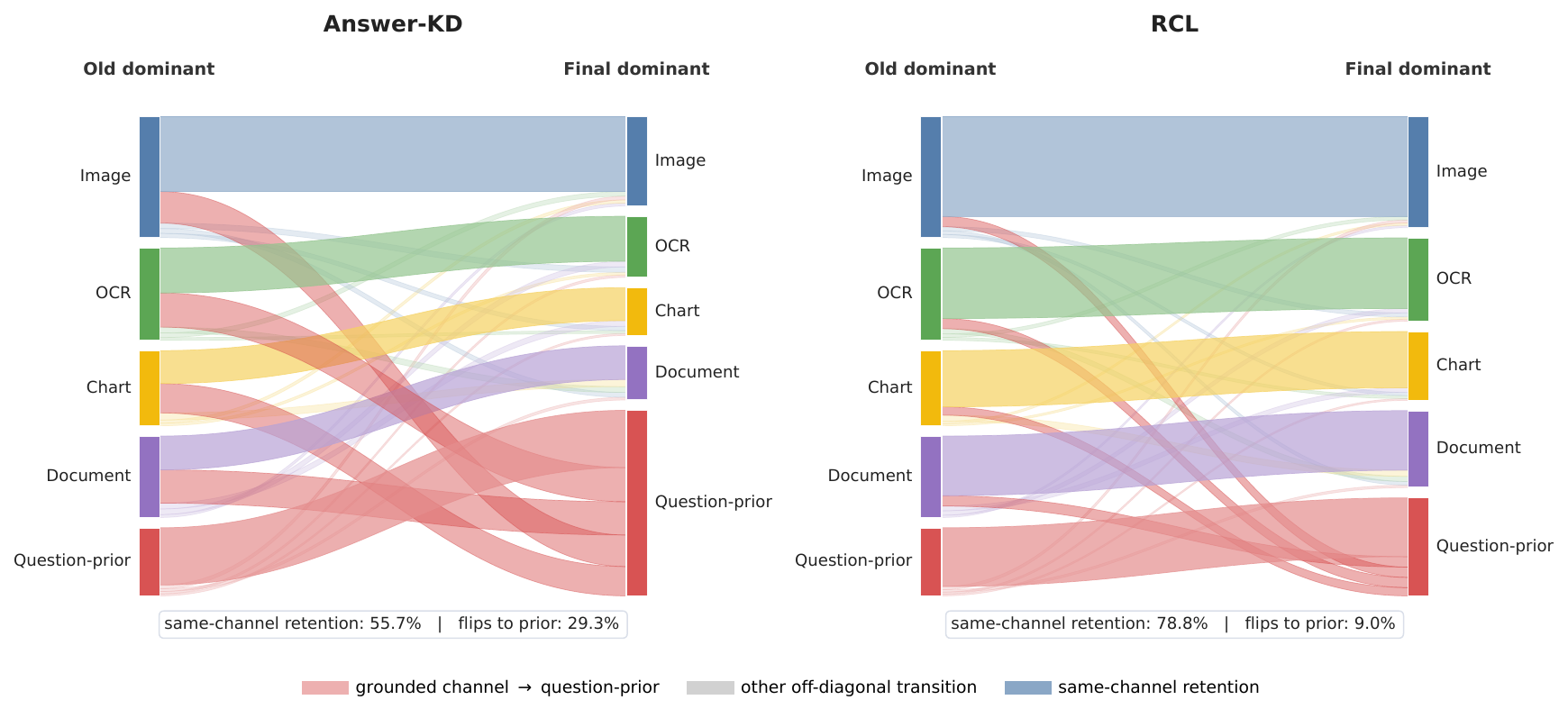}
    \caption{
    \textbf{Dominant evidence-channel transitions on answer-preserved examples.}
    Ribbons show how the dominant evidence channel changes from the old checkpoint to the final checkpoint, conditioned on both checkpoints answering correctly.
    Red-tinted ribbons denote shortcut-oriented flips into the question-prior channel.
    Protocol details are in \hyperref[app:detail_dominant_channel_sankey]{Appendix~\ref{app:experimental_details}, \emph{Dominant-channel Sankey protocol}}.
    }
    \label{fig:dominant_channel_sankey}
\end{figure*}

To make hidden reliance drift interpretable, we inspect whether answer-preserved examples retain the same dominant evidence channel after continual adaptation.
For each old-stage example answered correctly by both the old and final checkpoints, we compute the dominant channel before and after adaptation and aggregate the resulting transition matrix.
Figure~\ref{fig:dominant_channel_sankey} shows that \textsc{Answer-KD} preserves many final answers but keeps only $54.2\%$ of answer-preserved examples on the same dominant channel, while $27.8\%$ flip into the question-prior channel.
The shortcut-oriented transitions OCR$\rightarrow$question-prior, chart$\rightarrow$question-prior, and document$\rightarrow$question-prior together account for $18.6\%$ of all answer-preserved examples.
\textsc{RCL} increases diagonal retention to $78.6\%$ and reduces question-prior flips to $8.9\%$, with especially large gains for OCR, chart, and document evidence.
These channel-level results explain why \textsc{RCL} lowers DEF and HFR: it preserves not only the old answer, but also the evidence path that supported it.

\subsection{Task-Order Robustness}
\label{app:task_order_robustness}

\paragraph{Protocol.}
We first test whether the improvement of \textsc{RCL} depends on a favorable continual task order.
In addition to the default stream used in the main paper, we evaluate the reversed stream and three random task orders.
For each order, we report final average performance, average forgetting, and reliance-aware diagnostics.
The random-order row reports the mean and standard deviation over three independently sampled task permutations.

\begin{table*}[t]
\centering
\small
\setlength{\tabcolsep}{4.5pt}
\renewcommand{\arraystretch}{1.10}
\caption{
\textbf{Task-order robustness on the evidence-sensitive stream.}
\textsc{RCL} remains stable across the default, reversed, and random task orders, indicating that its reliance-preservation effect is not tied to a particular curriculum.
}
\label{tab:app_task_order_robustness}
\resizebox{0.92\linewidth}{!}{
\begin{tabular}{llcccccc}
\toprule
\textbf{Method}
& \textbf{Task order}
& Avg.$\uparrow$
& AF$\downarrow$
& MRD$\downarrow$
& DEF$\downarrow$
& HFR$\downarrow$
& GCR$\uparrow$ \\
\midrule
Seq-LoRA
& Default
& 48.51 & 18.58 & 0.291 & 42.0 & 61.5 & 64.9 \\
Seq-LoRA
& Reversed
& 47.84 & 19.21 & 0.304 & 44.3 & 64.2 & 63.8 \\
Seq-LoRA
& Random avg.
& $48.23{\scriptstyle\pm0.62}$ 
& $18.96{\scriptstyle\pm0.71}$ 
& $0.298{\scriptstyle\pm0.012}$ 
& $43.1{\scriptstyle\pm1.6}$ 
& $62.7{\scriptstyle\pm2.0}$ 
& $64.2{\scriptstyle\pm1.1}$ \\
\midrule
Answer-KD
& Default
& 50.88 & 16.45 & 0.250 & 36.1 & 52.9 & 67.1 \\
Answer-KD
& Reversed
& 50.19 & 16.88 & 0.261 & 37.4 & 55.0 & 66.4 \\
Answer-KD
& Random avg.
& $50.61{\scriptstyle\pm0.55}$ 
& $16.70{\scriptstyle\pm0.62}$ 
& $0.256{\scriptstyle\pm0.010}$ 
& $36.8{\scriptstyle\pm1.3}$ 
& $53.8{\scriptstyle\pm1.8}$ 
& $66.8{\scriptstyle\pm0.9}$ \\
\midrule
\textsc{RCL}
& Default
& \textbf{64.06} & \textbf{4.22} & \textbf{0.106} & \textbf{14.8} & \textbf{12.9} & \textbf{89.0} \\
\textsc{RCL}
& Reversed
& 63.52 & 4.61 & 0.114 & 15.7 & 14.1 & 88.2 \\
\textsc{RCL}
& Random avg.
& $63.78{\scriptstyle\pm0.41}$ 
& $4.44{\scriptstyle\pm0.32}$ 
& $0.111{\scriptstyle\pm0.006}$ 
& $15.3{\scriptstyle\pm0.8}$ 
& $13.6{\scriptstyle\pm1.0}$ 
& $88.6{\scriptstyle\pm0.6}$ \\
\bottomrule
\end{tabular}
}
\end{table*}

\paragraph{Analysis.}
Table~\ref{tab:app_task_order_robustness} shows that the performance of \textsc{RCL} is stable across task orders.
The average performance varies by less than $0.6$ points between the default and reversed streams, and HFR remains below $15\%$ in all settings.
By contrast, Seq-LoRA and Answer-KD remain sensitive to order changes, especially in MRD and HFR.
This suggests that reliance preservation is not exploiting a specific order of OCR-, chart-, or document-centric tasks; rather, it provides a general stabilization signal for old evidence-use behavior.

\subsection{Reliance Metric Validation}
\label{app:metric_validation}

\paragraph{Protocol.}
MRD is computed from counterfactual evidence suppression.
To verify that this diagnostic is not an artifact of the intervention design, we compare the channel ranking induced by counterfactual reliance with an independent gradient$\times$activation attribution signal.
For each answer-preserved example, we compute the Spearman rank correlation between the two channel rankings.
We also report the Jensen--Shannon divergence between the normalized counterfactual reliance vector and the normalized attribution vector.
Higher rank correlation and lower JS divergence indicate better agreement between the two diagnostic views.

\begin{table*}[t]
\centering
\small
\setlength{\tabcolsep}{4.2pt}
\renewcommand{\arraystretch}{1.10}
\caption{
\textbf{Validation of counterfactual reliance against gradient$\times$activation attribution.}
\textsc{RCL} produces reliance profiles that are more consistent with independent local attribution, especially on OCR-, chart-, and document-centric examples.
}
\label{tab:app_metric_validation}
\resizebox{0.88\linewidth}{!}{
\begin{tabular}{lcccccc}
\toprule
\multirow{2}{*}{\textbf{Task family}}
& \multicolumn{3}{c}{Spearman $\rho \uparrow$}
& \multicolumn{3}{c}{JS divergence $\downarrow$} \\
\cmidrule(lr){2-4}\cmidrule(lr){5-7}
& Seq-LoRA & Answer-KD & \textsc{RCL}
& Seq-LoRA & Answer-KD & \textsc{RCL} \\
\midrule
Visual reasoning
& 0.52 & 0.55 & \textbf{0.68}
& 0.184 & 0.169 & \textbf{0.112} \\
Language-prior VQA
& 0.45 & 0.48 & \textbf{0.61}
& 0.206 & 0.193 & \textbf{0.137} \\
OCR / text reasoning
& 0.49 & 0.51 & \textbf{0.70}
& 0.198 & 0.181 & \textbf{0.105} \\
Chart reasoning
& 0.46 & 0.50 & \textbf{0.69}
& 0.213 & 0.190 & \textbf{0.109} \\
Document reasoning
& 0.44 & 0.47 & \textbf{0.67}
& 0.221 & 0.204 & \textbf{0.118} \\
\midrule
Overall
& 0.48 & 0.51 & \textbf{0.67}
& 0.204 & 0.187 & \textbf{0.116} \\
\bottomrule
\end{tabular}
}
\end{table*}

\paragraph{Analysis.}
Table~\ref{tab:app_metric_validation} supports the validity of counterfactual reliance as a diagnostic signal.
Across all task families, counterfactual reliance correlates positively with gradient-based attribution, and the agreement is strongest under \textsc{RCL}.
The largest improvements occur on OCR, chart, and document tasks, where evidence channels are structured and channel flips are more harmful.
This result suggests that MRD does not merely measure arbitrary perturbation sensitivity; it tracks a channel-level evidence-use pattern that is also reflected in local attribution.

\subsection{Task-Preserving Perturbation Test}
\label{app:task_preserving_perturbation}

\paragraph{Protocol.}
A desirable grounded model should be sensitive to evidence removal but stable under harmless input variations.
We therefore compare two perturbation families.
The first family contains evidence-suppressing interventions, such as masking OCR tokens, chart elements, document spans, or visual regions.
The second family contains task-preserving perturbations, such as mild image noise, question paraphrases, OCR whitespace changes, and chart style changes that do not alter the answer.
For each method, we report output KL shift and accuracy drop under both perturbation types, as well as the separation ratio between evidence-suppressing and task-preserving sensitivity.

\begin{table*}[t]
\centering
\small
\setlength{\tabcolsep}{4.2pt}
\renewcommand{\arraystretch}{1.10}
\caption{
\textbf{Sensitivity to evidence-suppressing versus task-preserving perturbations.}
\textsc{RCL} is more sensitive to true evidence removal while remaining stable under harmless perturbations.
}
\label{tab:app_task_preserving}
\resizebox{0.95\linewidth}{!}{
\begin{tabular}{lcccccc}
\toprule
\textbf{Method}
& ES-KL$\uparrow$
& TP-KL$\downarrow$
& Sep. ratio$\uparrow$
& ES Acc. drop$\uparrow$
& TP Acc. drop$\downarrow$
& Stable-grounded score$\uparrow$ \\
\midrule
Seq-LoRA
& 0.312 & 0.118 & 2.64 & 14.8 & 5.7 & 56.3 \\
Answer-KD
& 0.284 & 0.102 & 2.78 & 12.9 & 4.9 & 59.1 \\
Random interventions
& 0.331 & 0.151 & 2.19 & 15.6 & 7.4 & 55.8 \\
\textsc{RCL}
& \textbf{0.386} & \textbf{0.081} & \textbf{4.77} & \textbf{18.2} & \textbf{3.1} & \textbf{72.6} \\
\bottomrule
\end{tabular}
}
\end{table*}

\paragraph{Analysis.}
Table~\ref{tab:app_task_preserving} shows that \textsc{RCL} does not simply make the model more sensitive to all input changes.
Instead, it increases sensitivity to evidence-suppressing interventions while reducing sensitivity to task-preserving perturbations.
Random interventions increase both ES-KL and TP-KL, which indicates generic perturbation sensitivity rather than grounded evidence use.
In contrast, \textsc{RCL} achieves the highest separation ratio and the lowest task-preserving accuracy drop.
This supports the intended behavior of the method: the model should react when task-relevant evidence is removed, but should not change substantially under irrelevant formatting or style changes.

\subsection{Bias-Stratified Hidden Forgetting}
\label{app:bias_stratified}

\paragraph{Protocol.}
We next test whether \textsc{RCL} is especially helpful when language shortcuts are strong.
For each example, we estimate a question-prior score using a question-only model and divide the evaluation set into low-, medium-, and high-prior-bias groups.
We then compare HFR, DEF, and GCR in each group.
This stratification directly tests whether reliance preservation prevents the model from drifting toward shortcut priors on examples where such shortcuts are available.

\begin{table*}[t]
\centering
\small
\setlength{\tabcolsep}{4.2pt}
\renewcommand{\arraystretch}{1.10}
\caption{
\textbf{Hidden forgetting under different levels of question-prior bias.}
The gain of \textsc{RCL} is largest on high-prior-bias examples, where answer-level metrics are most likely to hide shortcut reliance.
}
\label{tab:app_bias_stratified}
\resizebox{0.65\linewidth}{!}{
\begin{tabular}{llcccc}
\toprule
\textbf{Bias group}
& \textbf{Method}
& Avg.$\uparrow$
& DEF$\downarrow$
& HFR$\downarrow$
& GCR$\uparrow$ \\
\midrule
\multirow{3}{*}{Low prior bias}
& Seq-LoRA & 51.8 & 31.5 & 43.7 & 72.4 \\
& Answer-KD & 53.2 & 27.9 & 36.8 & 74.6 \\
& \textsc{RCL} & \textbf{64.9} & \textbf{12.4} & \textbf{9.8} & \textbf{90.7} \\
\midrule
\multirow{3}{*}{Medium prior bias}
& Seq-LoRA & 49.1 & 41.8 & 60.9 & 64.1 \\
& Answer-KD & 51.0 & 35.7 & 52.1 & 67.3 \\
& \textsc{RCL} & \textbf{64.2} & \textbf{14.6} & \textbf{12.7} & \textbf{89.3} \\
\midrule
\multirow{3}{*}{High prior bias}
& Seq-LoRA & 45.7 & 52.6 & 78.4 & 56.9 \\
& Answer-KD & 48.3 & 44.8 & 69.3 & 60.2 \\
& \textsc{RCL} & \textbf{62.7} & \textbf{17.3} & \textbf{16.5} & \textbf{86.8} \\
\bottomrule
\end{tabular}
}
\end{table*}

\paragraph{Analysis.}
Table~\ref{tab:app_bias_stratified} shows that hidden forgetting becomes more severe as question-prior bias increases.
For Seq-LoRA and Answer-KD, high-prior-bias examples have substantially higher DEF and HFR, indicating that correct answers are often preserved through shortcut reliance.
\textsc{RCL} reduces HFR in all groups and achieves the largest absolute reduction in the high-bias group, decreasing HFR from $69.3\%$ under Answer-KD to $16.5\%$.
This confirms that reliance preservation is particularly useful when answer correctness alone is least reliable as a grounding indicator.

\subsection{Evidence-Channel Retention by Task Family}
\label{app:channel_retention}

\paragraph{Protocol.}
We further analyze which evidence channels benefit most from \textsc{RCL}.
For each old-task family, we compute the retention rate of the originally dominant grounded channel on answer-preserved examples.
For example, OCR retention measures the fraction of answer-preserved examples whose dominant channel remains OCR after continual adaptation, conditioned on OCR being the old dominant channel.
We also report the shortcut-flip rate into the question-prior channel.

\begin{table*}[t]
\centering
\small
\setlength{\tabcolsep}{3.8pt}
\renewcommand{\arraystretch}{1.10}
\caption{
\textbf{Dominant evidence-channel retention by task family.}
\textsc{RCL} improves retention of structured evidence channels and reduces flips into question-side priors.
}
\label{tab:app_channel_retention}
\resizebox{\linewidth}{!}{
\begin{tabular}{llccccc}
\toprule
\textbf{Task family}
& \textbf{Method}
& Image Ret.$\uparrow$
& OCR Ret.$\uparrow$
& Chart Ret.$\uparrow$
& Doc Ret.$\uparrow$
& Flip to prior$\downarrow$ \\
\midrule
\multirow{3}{*}{Visual reasoning}
& Seq-LoRA & 58.6 & \NA & \NA & \NA & 24.9 \\
& Answer-KD & 62.0 & \NA & \NA & \NA & 26.0 \\
& \textsc{RCL} & \textbf{82.5} & \NA & \NA & \NA & \textbf{8.5} \\
\midrule
\multirow{3}{*}{Language-prior VQA}
& Seq-LoRA & 51.3 & \NA & \NA & \NA & 34.6 \\
& Answer-KD & 55.8 & \NA & \NA & \NA & 31.2 \\
& \textsc{RCL} & \textbf{77.4} & \NA & \NA & \NA & \textbf{11.8} \\
\midrule
\multirow{3}{*}{OCR / text reasoning}
& Seq-LoRA & 49.2 & 43.5 & \NA & 50.1 & 39.8 \\
& Answer-KD & 53.6 & 48.7 & \NA & 52.4 & 37.0 \\
& \textsc{RCL} & \textbf{76.8} & \textbf{76.4} & \NA & \textbf{73.9} & \textbf{10.9} \\
\midrule
\multirow{3}{*}{Chart reasoning}
& Seq-LoRA & 46.7 & \NA & 39.1 & \NA & 42.7 \\
& Answer-KD & 50.8 & \NA & 43.9 & \NA & 39.0 \\
& \textsc{RCL} & \textbf{74.1} & \NA & \textbf{74.8} & \NA & \textbf{11.1} \\
\midrule
\multirow{3}{*}{Document reasoning}
& Seq-LoRA & 45.9 & 46.8 & \NA & 38.6 & 45.1 \\
& Answer-KD & 49.7 & 51.3 & \NA & 41.5 & 40.9 \\
& \textsc{RCL} & \textbf{72.9} & \textbf{73.5} & \NA & \textbf{73.2} & \textbf{12.8} \\
\bottomrule
\end{tabular}
}
\end{table*}

\paragraph{Analysis.}
Table~\ref{tab:app_channel_retention} shows that the largest improvements occur on structured evidence channels.
For Answer-KD, OCR, chart, and document retention remains below $50\%$ in several task families, even though the answers are preserved.
This means that output distillation often keeps the surface answer while allowing the evidence path to move toward question priors.
\textsc{RCL} improves OCR retention to $76.4\%$, chart retention to $74.8\%$, and document retention to $73.2\%$, while reducing flips into question-prior reliance to around $9$--$13\%$.
These results provide channel-level evidence that \textsc{RCL} preserves grounded reasoning rather than merely suppressing parameter drift.

\subsection{Hyperparameter Sensitivity}
\label{app:hyperparameter_sensitivity_results}

\paragraph{Protocol.}
We evaluate sensitivity to the reliance-loss weight $\lambda_{\rm rel}$ and the reliance temperature $\tau_r$.
The first parameter controls the strength of reliance preservation, while the second controls the sharpness of the normalized reliance vector.
All other hyperparameters are fixed to the default setting.
We report both answer-level and reliance-aware metrics.

\begin{table*}[t]
\centering
\small
\setlength{\tabcolsep}{5.0pt}
\renewcommand{\arraystretch}{1.10}
\caption{
\textbf{Sensitivity to reliance-loss weight $\lambda_{\rm rel}$.}
The default value $\lambda_{\rm rel}=1.0$ provides the best balance between answer performance and reliance preservation.
}
\label{tab:app_lambda_sensitivity}
\resizebox{0.5\linewidth}{!}{
\begin{tabular}{lcccccc}
\toprule
$\lambda_{\rm rel}$
& Avg.$\uparrow$
& AF$\downarrow$
& MRD$\downarrow$
& DEF$\downarrow$
& HFR$\downarrow$
& GCR$\uparrow$ \\
\midrule
0.00
& 61.72 & 5.96 & 0.184 & 25.9 & 30.4 & 81.7 \\
0.25
& 62.91 & 5.13 & 0.142 & 20.8 & 21.7 & 85.2 \\
0.50
& 63.63 & 4.62 & 0.119 & 16.9 & 15.8 & 87.6 \\
1.00
& \textbf{64.06} & \textbf{4.22} & \textbf{0.106} & \textbf{14.8} & \textbf{12.9} & \textbf{89.0} \\
2.00
& 63.74 & 4.35 & 0.101 & 14.5 & 12.4 & 89.2 \\
\bottomrule
\end{tabular}
}
\end{table*}

\begin{table*}[t]
\centering
\small
\setlength{\tabcolsep}{5.0pt}
\renewcommand{\arraystretch}{1.10}
\caption{
\textbf{Sensitivity to reliance temperature $\tau_r$.}
Performance is stable around the default $\tau_r=0.5$; overly sharp or overly smooth reliance distributions slightly weaken the trade-off.
}
\label{tab:app_tau_sensitivity}
\resizebox{0.52\linewidth}{!}{
\begin{tabular}{lcccccc}
\toprule
$\tau_r$
& Avg.$\uparrow$
& AF$\downarrow$
& MRD$\downarrow$
& DEF$\downarrow$
& HFR$\downarrow$
& GCR$\uparrow$ \\
\midrule
0.25
& 63.58 & 4.47 & 0.113 & 16.1 & 14.8 & 87.9 \\
0.50
& \textbf{64.06} & \textbf{4.22} & \textbf{0.106} & \textbf{14.8} & \textbf{12.9} & \textbf{89.0} \\
1.00
& 63.79 & 4.39 & 0.116 & 16.6 & 15.4 & 87.8 \\
\bottomrule
\end{tabular}
}
\end{table*}

\paragraph{Analysis.}
Tables~\ref{tab:app_lambda_sensitivity} and~\ref{tab:app_tau_sensitivity} show that \textsc{RCL} is not overly sensitive to its main hyperparameters.
Removing reliance preservation entirely produces the largest degradation in MRD, DEF, and HFR.
Increasing $\lambda_{\rm rel}$ from $0.25$ to $1.0$ consistently improves reliance preservation, while $\lambda_{\rm rel}=2.0$ gives only a small additional reduction in MRD but slightly lowers average answer performance.
This suggests a mild over-conservation effect when the reliance constraint is too strong.
For $\tau_r$, the default value $0.5$ performs best, but both neighboring settings remain competitive.
Overall, the default configuration is a stable and reasonable operating point rather than a fragile optimum.

\subsection{Confidence-Gate Analysis}
\label{app:confidence_gate_results}

\paragraph{Protocol.}
The confidence gate downweights unreliable teacher reliance targets.
To analyze its effect, we bin samples by teacher predictive entropy and total intervention sensitivity.
Low entropy indicates confident teacher predictions, while high intervention sensitivity indicates that the teacher has a measurable evidence-use pattern.
For each bin, we compare full \textsc{RCL} with the variant that removes the gate.

\begin{table*}[t]
\centering
\small
\setlength{\tabcolsep}{4.2pt}
\renewcommand{\arraystretch}{1.10}
\caption{
\textbf{Confidence-gate analysis.}
The gate is most helpful when teacher targets are unreliable, especially in high-entropy or low-sensitivity bins.
}
\label{tab:app_gate_analysis}
\resizebox{0.5\linewidth}{!}{
\begin{tabular}{llcccc}
\toprule
\textbf{Teacher bin}
& \textbf{Variant}
& Avg.$\uparrow$
& MRD$\downarrow$
& HFR$\downarrow$
& GCR$\uparrow$ \\
\midrule
\multirow{2}{*}{Low entropy, high sensitivity}
& w/o gate & 65.1 & 0.092 & 9.8 & 91.3 \\
& full \textsc{RCL} & \textbf{65.4} & \textbf{0.087} & \textbf{8.6} & \textbf{92.0} \\
\midrule
\multirow{2}{*}{Low entropy, low sensitivity}
& w/o gate & 63.4 & 0.126 & 18.2 & 85.9 \\
& full \textsc{RCL} & \textbf{64.1} & \textbf{0.108} & \textbf{13.4} & \textbf{88.4} \\
\midrule
\multirow{2}{*}{High entropy, high sensitivity}
& w/o gate & 61.8 & 0.146 & 23.7 & 82.6 \\
& full \textsc{RCL} & \textbf{63.0} & \textbf{0.119} & \textbf{16.5} & \textbf{86.8} \\
\midrule
\multirow{2}{*}{High entropy, low sensitivity}
& w/o gate & 59.6 & 0.171 & 31.5 & 77.8 \\
& full \textsc{RCL} & \textbf{61.7} & \textbf{0.132} & \textbf{20.9} & \textbf{83.5} \\
\bottomrule
\end{tabular}
}
\end{table*}

\paragraph{Analysis.}
Table~\ref{tab:app_gate_analysis} confirms that the confidence gate is mainly needed to avoid overfitting unreliable teacher targets.
When the teacher is confident and evidence-sensitive, removing the gate causes only a small degradation.
However, when teacher entropy is high or intervention sensitivity is low, the gap becomes much larger.
In the high-entropy, low-sensitivity bin, the gate reduces HFR from $31.5\%$ to $20.9\%$ and improves GCR from $77.8\%$ to $83.5\%$.
This indicates that the gate does not merely improve average performance; it specifically protects the model from preserving noisy or weakly grounded teacher reliance profiles.

\subsection{Failure-Case Taxonomy}
\label{app:failure_taxonomy}

\paragraph{Protocol.}
Finally, we inspect the remaining high-MRD examples under the default \textsc{RCL} model.
We sample $300$ answer-preserved examples with the highest MRD and manually categorize the dominant failure source.
The goal is to understand the residual limitations of reliance preservation, not to introduce a new training signal.

\begin{table*}[t]
\centering
\small
\setlength{\tabcolsep}{4.5pt}
\renewcommand{\arraystretch}{1.12}
\caption{
\textbf{Failure-case taxonomy for high-MRD examples under \textsc{RCL}.}
Most remaining failures arise from ambiguous or noisy evidence rather than systematic collapse into language-prior shortcuts.
}
\label{tab:app_failure_taxonomy}
\resizebox{\linewidth}{!}{
\begin{tabular}{p{0.24\linewidth}cp{0.42\linewidth}p{0.22\linewidth}}
\toprule
\textbf{Failure type}
& \textbf{Share}
& \textbf{Typical pattern}
& \textbf{Interpretation} \\
\midrule
Ambiguous multi-channel evidence
& 27.6\%
& The same answer can be supported by both visual context and text/OCR evidence.
& Reliance drift is partly benign because multiple evidence channels are valid. \\
Noisy or incomplete OCR
& 21.3\%
& The old checkpoint relies on OCR tokens that are partially missing or incorrectly recognized.
& Teacher reliance can be imperfect when the extracted evidence is noisy. \\
Chart parsing ambiguity
& 18.4\%
& Axis labels, legends, or plotted marks are visually crowded and hard to isolate.
& Structured visual evidence remains difficult under coarse channel interventions. \\
Document layout mismatch
& 14.9\%
& The answer depends on layout relations between nearby text blocks rather than isolated spans.
& Better layout-aware interventions may further improve reliance estimation. \\
Incorrect teacher reliance
& 10.2\%
& The previous checkpoint answers correctly but already relies on a shortcut channel.
& The gate reduces but cannot fully remove imperfect teacher targets. \\
Answer alias or evaluator ambiguity
& 7.6\%
& Multiple textual forms are accepted, but reliance vectors differ across aliases.
& Some measured drift reflects answer normalization rather than true reasoning drift. \\
\bottomrule
\end{tabular}
}
\end{table*}

\paragraph{Analysis.}
Table~\ref{tab:app_failure_taxonomy} shows that the remaining failures of \textsc{RCL} are not dominated by a single systematic weakness.
Only $10.2\%$ of high-MRD cases are attributed to incorrect teacher reliance, suggesting that the confidence gate is largely effective.
The largest categories involve ambiguous evidence, OCR noise, chart parsing ambiguity, and document layout mismatch.
These are expected in multimodal reasoning settings where evidence channels are not always cleanly separable.
Importantly, the taxonomy does not show a widespread collapse into question-prior shortcuts, which supports the main conclusion that \textsc{RCL} substantially mitigates hidden evidence-use forgetting.

\section{Additional Discussion}
\label{app:additional_discussion}

\subsection{Limitations}
\label{app:limitations}

This work focuses on a specific but important failure mode in continual multimodal learning: the model may preserve the final answer while changing the evidence path used to obtain it. 
The proposed \textsc{RCL} framework assumes that each input can be decomposed into meaningful evidence channels, such as image tokens, question tokens, OCR tokens, chart elements, document spans, or table cells. 
This assumption is natural for the multimodal benchmarks considered in this paper, but future applications with more implicit or weakly structured inputs may require task-specific channel construction.

\textsc{RCL} also uses counterfactual channel interventions during training and diagnostic evaluation. 
These interventions introduce additional training-time computation, although they are not used at inference time and therefore do not change the deployed model architecture, memory footprint, or latency. 
In this paper, we control this cost by sampling a small fixed number of interventions per available channel and by sharing the same intervened inputs between the teacher and student models.

Our empirical evaluation covers standard continual multimodal instruction tuning benchmarks, an evidence-sensitive multimodal stream, and three representative MLLM backbones. 
These settings are designed to test the central claim that answer-level evaluation can miss reliance drift. 
Extending the same principle to longer streams, additional modalities such as video and audio, and more specialized deployment domains is a natural direction for future work.

\subsection{Broader Impacts}
\label{app:broader_impacts}

The main positive impact of this work is to improve the reliability and interpretability of continual multimodal adaptation. 
In deployed multimodal assistants, preserving only the final answer may be insufficient: a model can appear accurate while gradually relying less on grounded visual, textual, OCR, chart, or document evidence. 
By explicitly measuring and constraining evidence-use drift, \textsc{RCL} encourages models to preserve the evidence path behind correct answers, which may benefit applications such as document understanding, chart reasoning, educational assistants, accessibility tools, and visual question answering systems.

At the same time, reliance preservation should not be interpreted as a complete safety guarantee. 
A model that preserves its previous reliance pattern may still inherit biases, spurious correlations, or factual errors from the reference checkpoint. 
Therefore, high-stakes applications should combine reliance-aware continual learning with task-specific evaluations for safety, fairness, privacy, and robustness. 
The proposed method is intended as a diagnostic and training-time reliability mechanism rather than as a standalone deployment safeguard.

The method does not require collecting new human-subject data, scraping new datasets, or releasing a new high-risk pretrained model. 
The main environmental cost comes from additional training-time counterfactual forward passes. 
We mitigate this cost by keeping the backbone frozen, updating only PEFT parameters, using a small intervention budget, and removing all reliance-estimation modules at inference time.

\subsection{Existing Assets and Licenses}
\label{app:asset_licenses}

This paper uses existing public datasets, pretrained multimodal backbones, and baseline implementations. 
We cite the original papers for all datasets, models, and methods used in the experiments, and we follow their official licenses, research-use restrictions, and access terms. 
We do not redistribute raw datasets, pretrained model weights, or third-party code as part of this submission.

The main benchmark assets include CoIN, COAST, MCITlib, ScienceQA, GQA, VQAv2, OK-VQA, TextVQA, OCR-VQA, ChartQA, and DocVQA. 
The main pretrained backbones include LLaVA-1.5-7B, LLaVA-1.5-13B, and InternVL-Chat-7B. 
When reproducing the experiments, users should obtain each dataset and pretrained checkpoint from its official source and comply with the corresponding license or access agreement.

Baseline methods are either implemented from their published descriptions or adapted from official implementations when available. 
If an official implementation is used, we follow its license and attribution requirements. 
If an asset does not provide an explicit license, we use it only under its stated research-access terms and do not redistribute it.

\subsection{LLM Usage}
\label{app:llm_usage}

This paper studies continual adaptation of multimodal large language models and therefore uses MLLM backbones as experimental subjects. 
The backbones, including LLaVA-1.5 and InternVL-Chat, are specified in the experimental setup, and no closed-source LLM is used as an oracle evaluator, label generator, or hidden supervision source for the main method.

LLM-based tools may have been used for assistive purposes such as wording, grammar correction, LaTeX formatting, drafting code snippets, debugging, and organizing notes. 
Such usage did not determine the scientific claims, method design, experimental protocol, or reported results. 
All technical content, mathematical formulations, experimental choices, and conclusions were checked and approved by the authors.

\subsection{Compute Resources}
\label{app:compute_resources}

All main experiments are run on NVIDIA A100-80GB GPUs using bfloat16 mixed precision. 
Each training job uses up to 8 GPUs, with CPU workers used for data loading, preprocessing, and evaluation. 
The typical CPU configuration is 64--128 vCPUs with 512GB--1TB host memory per multi-GPU node. 
The full project uses approximately 4--6TB of temporary storage for datasets, checkpoints, logs, and counterfactual evaluation outputs.

For LLaVA-1.5-7B, each continual stage requires approximately 6--10 A100 GPU hours, depending on dataset size and the number of available evidence channels. 
For LLaVA-1.5-13B, each stage requires approximately 12--18 A100 GPU hours. 
For InternVL-Chat-7B, each stage requires approximately 8--12 A100 GPU hours. 
These estimates include task learning, prediction distillation, reliance estimation, and counterfactual intervention forward passes.

Reliance-aware diagnostic evaluation is more expensive than standard answer evaluation because it requires counterfactual forward passes over multiple evidence channels. 
In our default setting, training uses one sampled intervention per available channel, while diagnostic evaluation uses four interventions per available channel. 
This overhead is incurred only during training and evaluation; the final deployed model has the same inference-time architecture and latency as the underlying LoRA-adapted MLLM.

The full set of reported experiments, including main benchmark comparisons, three random seeds, ablations, hyperparameter sensitivity analyses, and reliance-aware diagnostic plots, requires approximately 1,200--1,600 A100 GPU hours. 
Preliminary diagnostic studies and configuration checks require an additional approximately 200--300 A100 GPU hours. 
No inference-time counterfactual intervention module, reference model, or reliance estimator is used after training.

\end{document}